\documentclass[10pt, conference]{IEEEtran}
\usepackage{enumerate}
\usepackage{booktabs}
\usepackage{multirow}
\usepackage{algorithm}
\usepackage{algpseudocode}
\usepackage{comment}
\usepackage{amsmath,amssymb}
\usepackage{latexsym}
\usepackage{graphicx}
\usepackage{subfigure}
\usepackage{epstopdf}
\usepackage{epsfig}
\usepackage{xspace} 
\usepackage[bottom]{footmisc}
\usepackage[11pt]{moresize}
\usepackage{url}
\begin{document}
\title{PyGFI: Analyzing and Enhancing Robustness of Graph Neural Networks Against Hardware Errors}
\author{
	Ruixuan Wang$^{*}$, Fred Lin$^{\ddag}$, Daniel Moore$^{\ddag}$, Sriram Sankar$^{\ddag}$, Xun Jiao$^{*}$$^{\ddag}$   \\
        $^{*}$Villanova University,
        $^{\ddag}$Meta Platforms, Inc. \\
	\{rwang8, xun.jiao\}@villanova.edu, \{fanlin, danielmoore, sriramsankar\}@meta.com \\
	%\vspace{-0.5cm}
}
\maketitle
\begin{abstract}
Graph neural networks (GNNs) have emerged as a promising deep learning (DL) paradigm for analyzing graph-structured data, showing success in various domains such as recommendation systems, social networks, and electronic design automation (EDA). Despite their growing popularity and deployment on modern hardware systems, including dedicated accelerators, the fault tolerance and robustness of GNNs have been generally overlooked. This is particularly concerning given the increasing prevalence of hardware errors in large-scale fleet systems where GNNs are being deployed.
As the workload of large-scale AI tasks continues to grow, modern data center fleets have become more heterogeneous and scaled with an increasing number of computing accelerators (e.g., GPUs). Consequently, this significantly increases the risk of hardware failures, which can ultimately result in wrong results or degraded service. Therefore, it is critically important to study the robustness of widely deployed AI models and improve their fault tolerance and resilience by conducting error injection campaigns.
This paper presents the first study of GNN robustness against hardware errors. The study aims to understand the relationship between hardware errors and GNN performance and develop new opportunities for future GNN accelerator design and architectural optimization. We develop PyGFI, the first versatile and scalable error injection framework for GNN on top of the widely-used DL frameworks. Using this framework, we conduct an extensive error injection campaign on various GNNs models and application datasets. The study finds that the robustness of GNNs models varies significantly across different models and application datasets. We propose a topology-aware node-level error protection method as well as explore existing error protection methods to enhance GNN robustness, which can improve the robustness of GNNs by orders of magnitude. This study highlights the importance of addressing hardware errors in GNNs and provides insights into the factors affecting GNN robustness, which can inform future GNN accelerator design, architectural optimization, and large-scale fleet management. 
\end{abstract}
% \small
% \keywords{Graph neural networks, silent data corruption, error
\section{Introduction}
\label{sec:intro} 
In recent years, there has been a growing interest in graph neural networks (GNNs) due to their exceptional performance in learning from graph-structured data, which is prevalent in various applications such as recommendation systems~\cite{wu2020graph}, social networks~\cite{fan2019graph}, and electronic design automation (EDA)~\cite{lopera2021survey}. GNNs have been implemented on diverse hardware platforms and specialized accelerators such as GPU~\cite{liu2020g3}, ASIC~\cite{kiningham2022grip}, and FPGA~\cite{li2022hyperscale}. However, despite the growing development and implementation of GNNs on hardware, the impact of hardware errors on GNN robustness has not been explored systematically. 

As large-scale AI tasks continue to grow, modern data center fleets have become more heterogeneous and scaled with an increasing number of computing accelerators, including powerful GPUs. Meanwhile, hardware systems are advancing toward the deep nanometer regime. These developments have significantly increased the risk of hardware failures and data corruption, which can lead to mis-predicted results. It is therefore imperative to address the fault tolerance and resilience of AI systems and ensure their continued effectiveness in the face of these challenges. Common types of hardware errors include radiation-induced soft errors~\cite{hazucha2003neutron}, variation-induced timing errors~\cite{ernst2004razor}, as well as errors caused by intentional design choices such as approximate computing~\cite{han2013approximate} and voltage scaling~\cite{zhang2018thundervolt}. These errors usually result in bit flips during computation, leading to system corruptions, permanent errors, or silent data corruption (SDC), which have been observed in the large-scale modern infrastructure of major corporations such as Meta~\cite{dixit2022detecting} and Google~\cite{bacon2022detection}.

For example, SDC is a type of fault that occurs when data is corrupted in a way that is not detected by the system's error detection mechanisms, resulting in erroneous results that are often not apparent to the user. The impact of SDC on computing systems and application services can be severe, as it can lead to inaccurate or incorrect results that can have significant consequences. Tech companies such as Google~\cite{bacon2022detection} and Meta~\cite{dixit2022detecting} have reported instances of SDC. For example, we have detected SDC in our large-scale fleet, with hundreds of CPUs detected among hundreds of thousands of machines. In the same year, Google published a study in which they analyzed data from their production systems and found that SDC occurred more frequently than previously believed, with an average rate of ``a few per several thousands of machines''.

Conventional computing typically requires negligible error rates (e.g., $< 10^{-15}$~\cite{jesd2182010solid}) during computation to guarantee correct execution. Such a requirement can poses a high design cost at device and architecture levels as hardware designers need to add extra redundancy or protection mechanisms to hardware, or incorporate error detection and correction mechanisms at various levels of the system architecture, from the memory and storage subsystems to the processor and interconnect fabric. To combat this challenge, designers are devising a scenario to continue the operations even in presence of errors, leading to approximate computing which allows occasional errors in a system. Specifically, recent research has shown that AI/ML models such as convolutional neural networks (CNNs) can exhibit a certain level of resilience or robustness to hardware errors, particularly in the presence of noisy or incomplete data~\cite{jiao2017assessment, zhang2018thundervolt}. These techniques can be leveraged to design more efficient hardware~\cite{reagen2016minerva, zhang2018thundervolt}, by reducing the precision or accuracy requirements of the computations and thereby reducing the energy consumption and hardware overhead.

Given the growing interest in GNNs as a popular AI/ML model, this paper is inspired by the aforementioned phenomenon to answer the following question: Are GNNs models robust to hardware errors/faults like we have seen in other AI/ML models? To this end, we systematically assess the impact of hardware errors on the GNNs model accuracy. Such assessment of GNN robustness can be broadly applied across a range of devices and architectures running GNN, which is vital as GNNs can run on GPUs, ASIC, or other accelerators, and the optimal architecture for GNN system is still widely under debate. Our results show that the robustness of GNN varies significantly across models and application datasets.

Specifically, we make the following major contributions: 
\begin{itemize}
    \item To the best of our knowledge, we conduct the first study of GNN robustness to hardware errors. We perform extensive error injection campaigns on four popular GNNs models under three datasets by sweeping a wide range of error rates to assess the impact of hardware errors on GNN accuracy. We observe that the robustness of GNN varies by orders of magnitude with respect to different models and application datasets. 
    \item We develop PyGFI, the first versatile and scalable error injection framework for GNN on top of the widely-used PyTorch and PyTorch Geometric libraries. PyGFI is able to inject large-scale bit-flips to the GNNs model with user-defined error rates, injection sites (e.g., specific layer or node), and injection data (e.g., weights or activation output).
    \item We propose an error mitigation approach to enhance the robustness of GNNs. The key idea is by leveraging the unique characteristic of GNN, message passing, and for each node, we automatically filter the extreme values by leveraging passed messages from neighbors. Experimental results show that we can enhance the GNN robustness by orders of magnitude without modifying any hardware design. 
\end{itemize}

\section{Background on GNN}
%1 page
Graph neural networks (GNNs) are a type of neural network that can process and analyze graph-structured data. In contrast to traditional neural networks that operate on grid-structured data, such as images or speech signals, GNNs can handle data represented as graphs, such as social networks, recommendation systems, or hardware circuits. GNNs can learn the structural and semantic information of the graph, such as the connectivity between nodes, node attributes, and edge weights, to perform various tasks, including node classification, link prediction, and graph classification. GNNs typically involve message passing between nodes, where each node aggregates information from its neighbors and updates its own representation based on this information. 

Recently, GNNs have demonstrated ground-breaking performances on many graph-related tasks, such as recommendation systems~\cite{wu2020graph}, social networks~\cite{fan2019graph}, and electronic design automation (EDA)~\cite{lopera2021survey}. For example, GNNs can be used to model user-item interactions in recommendation systems, where the user-item interactions are represented as a graph~\cite{wu2020graph}. GNNs can learn the latent features of items and users based on their interactions, and use this information to make personalized recommendations. In the EDA domain, GNNs have been used to optimize the design of electronic circuits, where the circuits are represented as graphs. GNNs can learn the functional relationships between components, and use this information to improve the performance and efficiency of the circuits~\cite{lopera2021survey}. 

The input to GNNs are graphs, which are typically defined as $G = (V, E)$, where $V$ is the set of nodes and $E$ is the set of edges. Graphs can be categorized as directed/undirected graphs (depending on if edges are directed/undirected), homogeneous/heterogeneous graphs (depending on if nodes are of one/multiple types), and hypergraphs (an edge can join any number of nodes). 
The main idea of GNNs is a process called message passing, where GNNs iteratively aggregate feature information from neighbor nodes and use it to update the information for the current node. During the aggregation process, GNNs can simply treat each neighbor equally with the mean-pooling operation, or treat them differently with the attention mechanism.
During the update step, GNNs integrate the aggregation information with the information of the current node. Various strategies are proposed to integrate the two feature information, such as GRU mechanism~\cite{li2015gated}, concatenation with nonlinear transformation~\cite{hamilton2017inductive} and sum operation~\cite{velivckovic2017graph}. 
Some representative GNNs include graph convolutional networks (GCN)~\cite{kipf2016semi} and graph attention networks (GAT)~\cite{velivckovic2017graph}. 

%\vspace{-0.5cm}
\begin{equation}
\small
    \begin{aligned}
        h_{v}^{m+1} = RELU(W \cdot AGG(h_{v}^{m}, u \in \{N(v) \cup \{v\}\}))
    \end{aligned}
\label{eqn:gcn}
\end{equation}

\begin{figure}[htbp!]
    \centering
    \includegraphics[width=0.75\columnwidth]{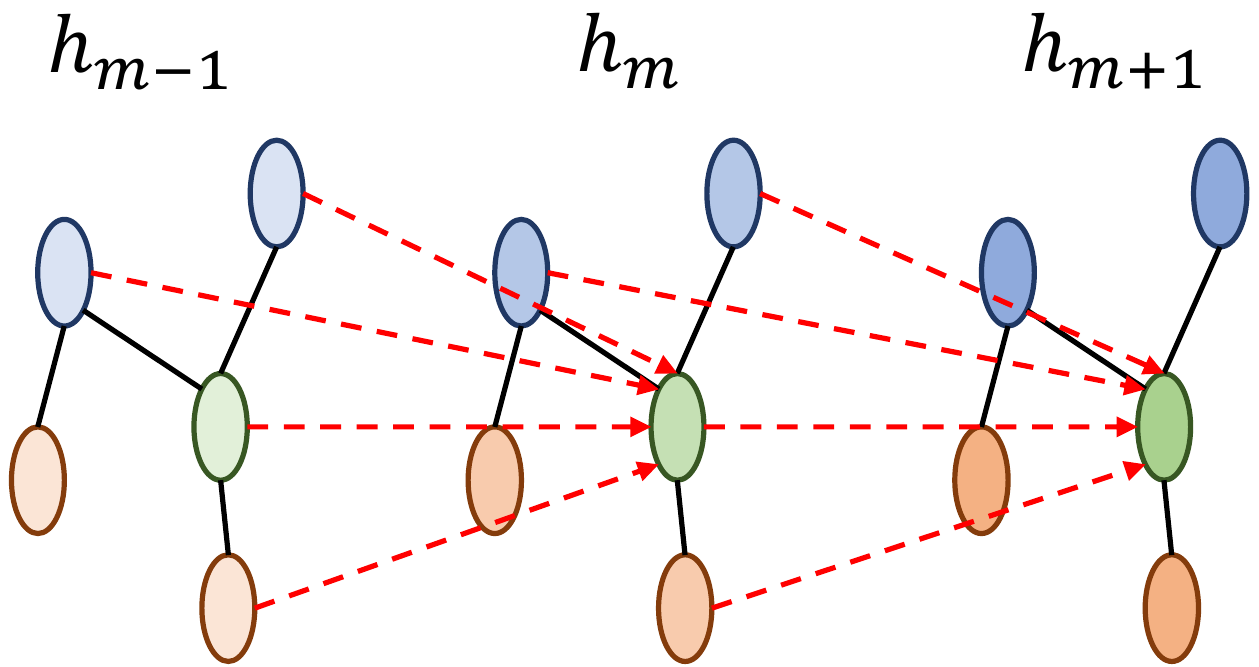}

    \caption{A general indication of the data flow of node representations in GCN model, each node receives information from the node itself (green node) and its neighbors (blue and red nodes) in the previous layer and computes the new representation in the next layer.}
    % \vspace{-0.5cm}
    \label{fig:gcn1}
\end{figure}

\begin{figure}[htbp!]
    \centering
    \includegraphics[width=0.75\columnwidth]{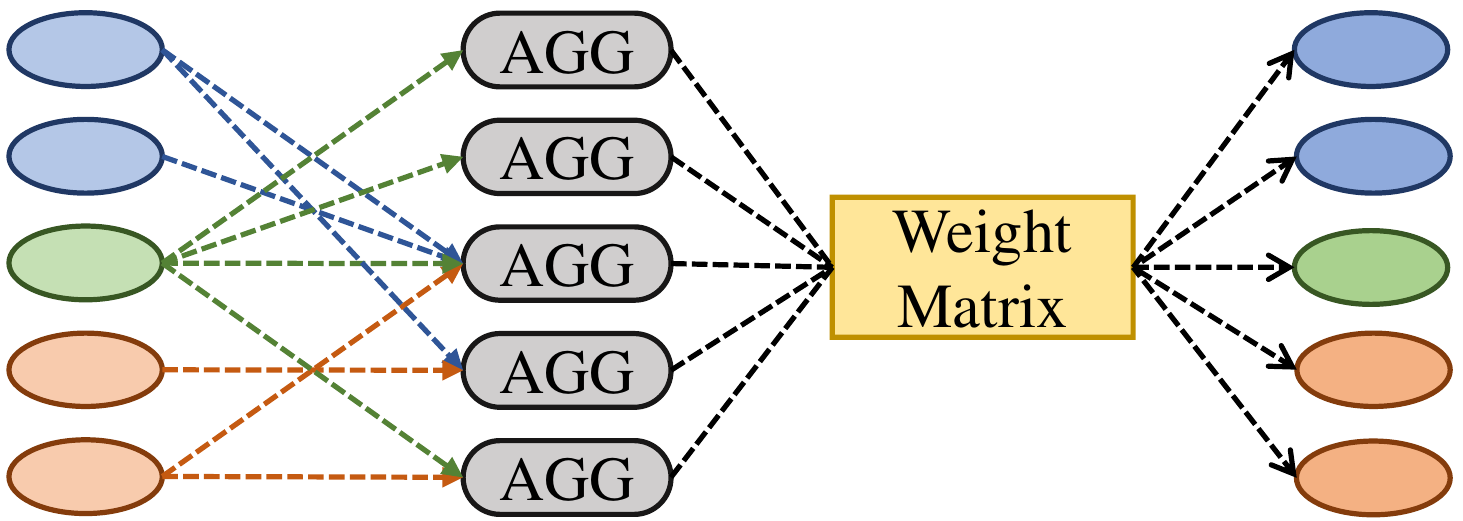}
    
    \caption{The representation of the message passing process in GCN model, where $AGG$ indicates the mean-aggregation operation between node $v$ and its neighbors $N(v)$}
    % \vspace{-0.5cm}
    \label{fig:gcn2}
\end{figure}

Here we use GCN as an example to illustrate the general process of GNNs. GCN approximates the first-order eigendecomposition of the graph Laplacian to iteratively aggregate information from neighbor nodes. Specifically, Considering a node $v$ in graph $G = (V, E)$, it updates the node embedding by Eq.~\ref{eqn:gcn}, where $h_{v}^{m}$ and $h_{v}^{m+1}$ are the representation of this node at layer $m$ and $m+1$ in GCN. From the GCN perspective, $AGG$ and $N(v)$ are the representation aggregation and the neighbors of node $v$ respectively. Fig.~\ref{fig:gcn1} and Fig.~\ref{fig:gcn2} illustrate the data flow and the process of message passing in GCN.

\section{PyGFI: Error Injection Tool for GNNs}
%1 page
In this section, we present a novel error injection framework \textbf{PyGFI}, especially for the error analysis for GNNs models. Although the existing research tools~\cite{reagen2018ares, chen2020tensorfi, mahmoud2020pytorchfi} can inject errors to CNN/RNN models, they cannot inject errors to GNNs models because of the nature of different architecture and variant parameters passed in the GNNs layer. One main difference between GNNs and CNN models is that the CNN layer takes a single input signal and applies convolution operations, while GNN layers utilize both node representation and the corresponding adjacency matrix of the graph for message passing. The CNN error injector lack the interfaces or ports to access the topological information without redesigning the architecture. Thus, existing CNN error injectors are naturally incompatible with the GNNs models.

\begin{figure}[htbp!]
	\centering
    % \hspace{-8pt}
	\includegraphics[width=0.28\textwidth]{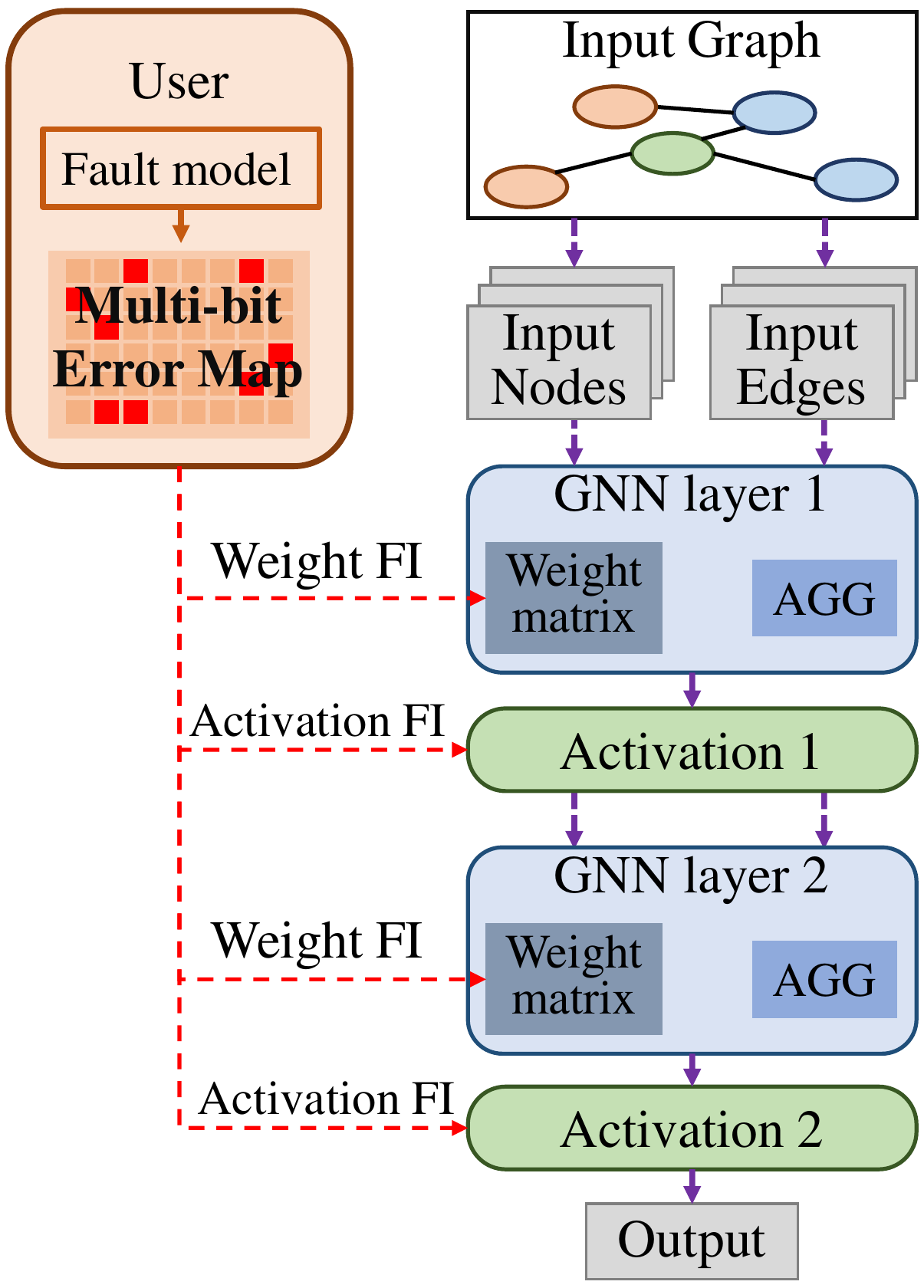}
	\caption{The error injection pipeline of PyGFI framework on a 2-layer GNNs model. PyGFI takes the user-customized fault model and the user-defined multiple-bit flipping error maps as the input to inject weight and activation errors.}
	\label{fig:pygfi}
    % \vspace{-0.4cm}
\end{figure}

In this section, we describe the hardware fault models used in this work and the development flow of PyGFI. An overview of PyGFI is illustrated in Figure~\ref{fig:pygfi}. PyGFI is a runtime error injection tool designed for GNNs models built on top of PyTorch~\cite{paszke2019pytorch} and PyTorch Geometric (PyG) library~\cite{fey2019fast}. PyG is an open-source library built upon PyTorch for training GNNs so it inherits/shares most interfaces with PyTorch. The primary purpose of PyG is to simplify the development and training of GNNs for various applications related to structured data. PyG library offers a comprehensive set of tools for DL on graphs and other irregular structures, collectively referred to as geometric DL.

\subsection{Error Model}
As the transistor size scales down to deep nanometer era, silent errors in microelectronic circuits are increasingly prevalent, caused by manufacturing process variation, voltage noise, and temperature~\cite{gupta2012underdesigned}. Hardware designers are forced to use worst-case design by adding safety margins such as slowing down clock frequency, increasing supply voltage, and/or add redundancy in design (e.g., ECC).
%However, this practice leads to overly conservative design as the margins often exceed 40\% of the nominal target specifications~\cite{gupta2019variability}.
While there are intensive efforts to reduce these wasteful/costly margins by reclaiming performance or energy efficiency, they risk the occurrence of errors/errors under certain operating conditions~\cite{gupta2012underdesigned}. For example, SRAM, DRAM, and flash memory cells have all been demonstrated susceptible to variation-caused errors and soft errors caused by cosmic particle strikes due to their dense circuitry, reduced supply voltage, and bit encoding with few electrons~\cite{gupta2012underdesigned}, that can lead to SDCs in memory and computation paths.

In this paper, we follow the same practice of existing DL robustness studies~\cite{reagen2018ares, li2017understanding, kim2018matic} by focusing on the errors in memory, which are most relevant to DL/GNN systems because of their demanding storage requirements of parameters including weights and intermediate results (e.g., activation outputs). We consider the use of the most widely-used \textit{bit flip} fault model~\cite{li2017understanding, reagen2018ares}, where each bit will have a certain probability to be flipped, i.e., bit error rate (BER). This is consistent with the practical scenario where a fault occurs in the memory that is not caught by, e.g., ECC. The bit flips will have a direct impact on the value of weights or intermediate results, e.g., activation output. With that being said, PyGFI can also be used to inject computation parts and datapath errors, e.g., multiply-accumulate (MAC) units, because the input and output of MAC units are weights and activation outputs that are stored in memory. Therefore, to inject MAC-related errors, we can manipulate MAC input/output stored in memory, which are essentially activation outputs that are supported by PyGFI. Further, PyGFI would be able to accept any customized error models, which will be explained below.

\subsection{Error Injection Design}

To mimic the actual scenario of the bit-flip error in the real-world computational infrastructure, we propose PyGFI, and present several error injection principles in it. 

\textbf{Multi-bit Error Injection: }
During our observation and experiments on a large-scale fleet, we find that one operation/instruction can have multiple bit flips. Similar circumstances are also discovered in the earlier works~\cite{sangchoolie2020empirical,cho2013quantitative} that soft errors are manifested as multiple bit-flips in modern devices. Thus, we consider the multi-bit flip as our fault model. Implementation of the (multiple) bit flip model with each bit having a certain probability to flip is non-trivial, especially considering the extremely large number of bits and parameters in a GNNs model. The most straightforward way of implementing the bit-level flip model is to iterate through each bit and generate a random number, which will be compared against the pre-defined BER (i.e., bit flip probability) to determine if that bit will get flipped or not. However, this way of implementing bit flip is extremely time-consuming due to the size of parameters in a GNN. To resolve this issue, we use a pre-configured error map as below to inject errors. 

\textbf{Error Map-based Injection: }
To efficiently inject errors to millions of GNN parameter bits, for each experiment (defined as performing one round of inference across all testing data), we first generate an error injection map based on the corresponding GNNs model and data type, with each bit position assigned flip/not-flip with a certain probability passed as a parameter to the function. This probability corresponds to the predetermined bit error rate (BER). Using this method, we can quickly configure the bits that will get flipped and the error injection process can speed up by 15-50X.

\subsection{Error Injection on GNNs models}

In this paper, we recognize the errors in different locations in the GNNs model: model weights, and activation outputs. Bit-flip errors in different positions require different error injection implementations.

\subsubsection{Error Injection on Weight Matrices}
Injecting errors into the weights of GNNs models can be implemented by adjusting the values in weight matrices. Since we focus on the bit error robustness during the inference phase and utilize the pre-trained GNNs models, we can change the values in the weight matrices based on the corresponding error map generated according to the specified BER, where we point to the position and flip the related bit(s) based on the generated error map.

%  \begin{figure}[htbp!]
%  	\centering
%      % \hspace{-20pt}
%  	\includegraphics[width=0.5\textwidth]{./figures/weight_error.pdf}
%  	\caption{GNNs model weights error injection based on customized function and error map.}
%   % Currently, PyGFI support single/multi-bit flip and random stuck at fault injection.}
% \label{fig:code1}
%  \end{figure}

\subsubsection{Error Injection on Activation Outputs}
Injecting errors to activation output need to happen during runtime because the value after bit flips are input-dependent, e.g., activation outputs are only visible during runtime. PyGFI allows the injection of errors on weight values as well as activation output, which can be used for perturbation analysis such as (resilient/efficient) GNN algorithm/hardware design, adversarial/poison attacks, interpretability, and explainability.

One of the challenges of error injection during the runtime on DL models is the modifications of on-the-fly activation outputs. This modification is non-trivial because it either (1) needs the user to add an additional layer dedicated for perturbation, which adds extra burden on users or (2) needs to manually modify the low-level DL framework utilities to intercept the activation output which compromises the scalability and portability of the DL framework subject to future maintenance and patches. To address the issue, we develop an activation error injection flow based on the \textit{hook} interface of PyTorch, where we use the hook interface to inject bit-flips in the activation outputs. hook has been used in visualizing/modifying internal structures of DL models~\cite{pawade2021xai, mahmoud2020pytorchfi} and can be attached to an arbitrary basic module in a neural network with a format of \textit{hook(module, input, output)}. The \textit{module} refers to a PyTorch neural network module such as a layer or a block, and \textit{input, output} refers to the input/output of this module. Note that when a  \textit{hook($\cdot$)} is registered to a neural network, the \textit{hook($\cdot$)} will automatically execute the corresponding functionalities, such as obtaining and modifying the input and output of the module, during the module runtime.

Consequently, if we want to perturb the input/output of a given layer, we will perform the following two steps: 
\begin{itemize}
    \item Step 1: We define an \textit{Injection($\cdot$)} function, where we can manually change or perturb the values of input/output of the layer. Note that we can define a customized error model here in the \textit{Injection($\cdot$)} function. 
    \item Step 2: We implement the \textit{Injection($\cdot$)} through \textit{hook($\cdot$)} function on the targeted layers. This finishes the attachment of the \textit{hook($\cdot$)} to the module, and whenever the module is executed, the \textit{hook($\cdot$)} will also be executed to perform its functionality. 
\end{itemize}

\begin{algorithm}[htbp!]
\small
    \caption{Error Injection Example on Target Layers}
    \algrenewcommand\algorithmicrequire{\textbf{Input}}
    \algrenewcommand\algorithmicensure{\textbf{Output}}
    \begin{algorithmic}[1]
    \Require Target model $M$, Target layer list $L$
    \Require User-defined Error Map $U$
    \Ensure Error Injected Model $M'$
    \State $M'$ $\leftarrow$ copy($M$)
    \For{$Layer$ in $M$} \Comment{Traverse all the layers in the model}
        \If{$Layer$ in $L$} \Comment{Find the target layers}
            \State hook $\leftarrow$ Injection($U$, BITFLIP)  \Comment{Define activation hook}
            \State $Layer$ $\leftarrow$ hook($Layer$, $input$, $output$) \Comment{Register hook}
        \EndIf
    \EndFor
    \State Execute $M'(Node, Edge)$\Comment{Activate all registered hooks} \\
    \Return $M'$ \Comment{$M'$ is the error injected model}
    \end{algorithmic}
    \label{alg:pygfi}
\end{algorithm}

We use a simple example indicated in Alg.\ref{alg:pygfi} for error injection based on PyGFI. Note that here, since we are only interested in assessing the error impact on inference accuracy, we will only do \textit{register\_module\_forward\_hook} where the \textit{hook($\cdot$)} will only be executed during the forward process.

\section{Error Mitigation Method}
\label{sec:mitigation}
In this section, we describe how we mitigate the impact of hardware errors in GNNs models by proposing a GNN topology-aware node-level error protection method in Section~\ref{sec:filtering} as well as exploring two existing protection methods in Section~\ref{sec:mask} and Section~\ref{sec:clip}.

\subsection{Error Masking}
\label{sec:mask}
% Preliminary studies have illustrated that the ECC can be deployed to detect and correct bit errors. In this paper, w
We explore the Razor double-sampling-based circuitry, which is a low-cost error masking technique to detect and mask errors~\cite{das2006self} caused by timing violations. Unlike parity bits protection~\cite{jahinuzzaman2009design} that can also detect memory fault but provides no information on the location of the affected bit, the Razor-based method can detect erroneous bits and their locations. The detailed implementations of Razor detection circuitry are out of the scope of this study and can be found in the related work~\cite{das2006self, reagen2016minerva}. 

Here we apply the masking error mitigation approaches and implement two masking methods in different granularities, including word-level and bit-level, respectively, that have been used in CNN error mitigation study~\cite{reagen2016minerva}. In particular, for word masking, upon detecting any bit flips, it sets the entire word to 0. While for bit-level masking, it only recovers the faulty bit(s) to 0. The experimental results will be discussed in Sec.~\ref{sec:exp}.

\subsection{Weight and Activation Clipping}
\label{sec:clip}
%Considering the flexibility and efficiency of error mitigation, we further present the software-based mitigation approach on weight and activation outputs, without hardware-level modification. 
Previous works have studied the refinement of model fault tolerance through weight and activation clipping~\cite{hoang2020ft, arjovsky2017wasserstein}. Since the bit-flip error will perturb the parameter value distribution from the original distribution, we can set a threshold to filter out the outliers and decrease the influence of bit-flip errors. 
The clipping function is denoted in Eq.~\ref{eqn:clip}, where the ceiling ($C$) and floor ($F$) values are the pre-defined threshold for clipping and identical for all the GNNs models across our experiments, while the values of weight parameters ($x$) are clipped into the range of $[F, C]$. 

%\vspace{-0.5cm}
\begin{equation}
% \small
    \begin{aligned}
        Clip(x, C, F) = max(F, min (x, C))
    \end{aligned}
\label{eqn:clip}
\end{equation}

\begin{figure}[htbp!]
    \centering
    \subfigure[GCN]{
        \includegraphics[width=0.46\columnwidth]
        {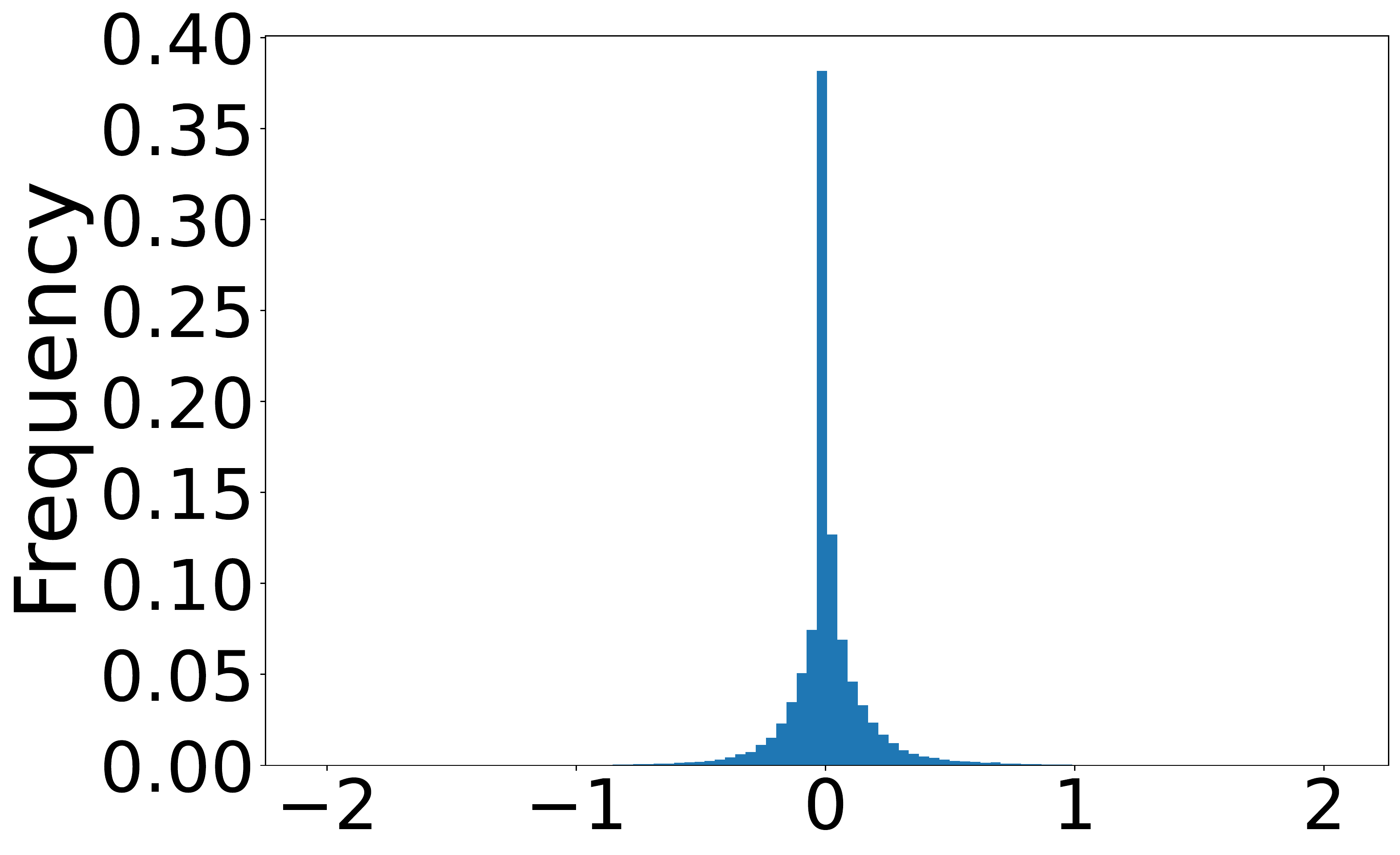}
    }
    \subfigure[GAT]{
        \includegraphics[width=0.46\columnwidth]
        {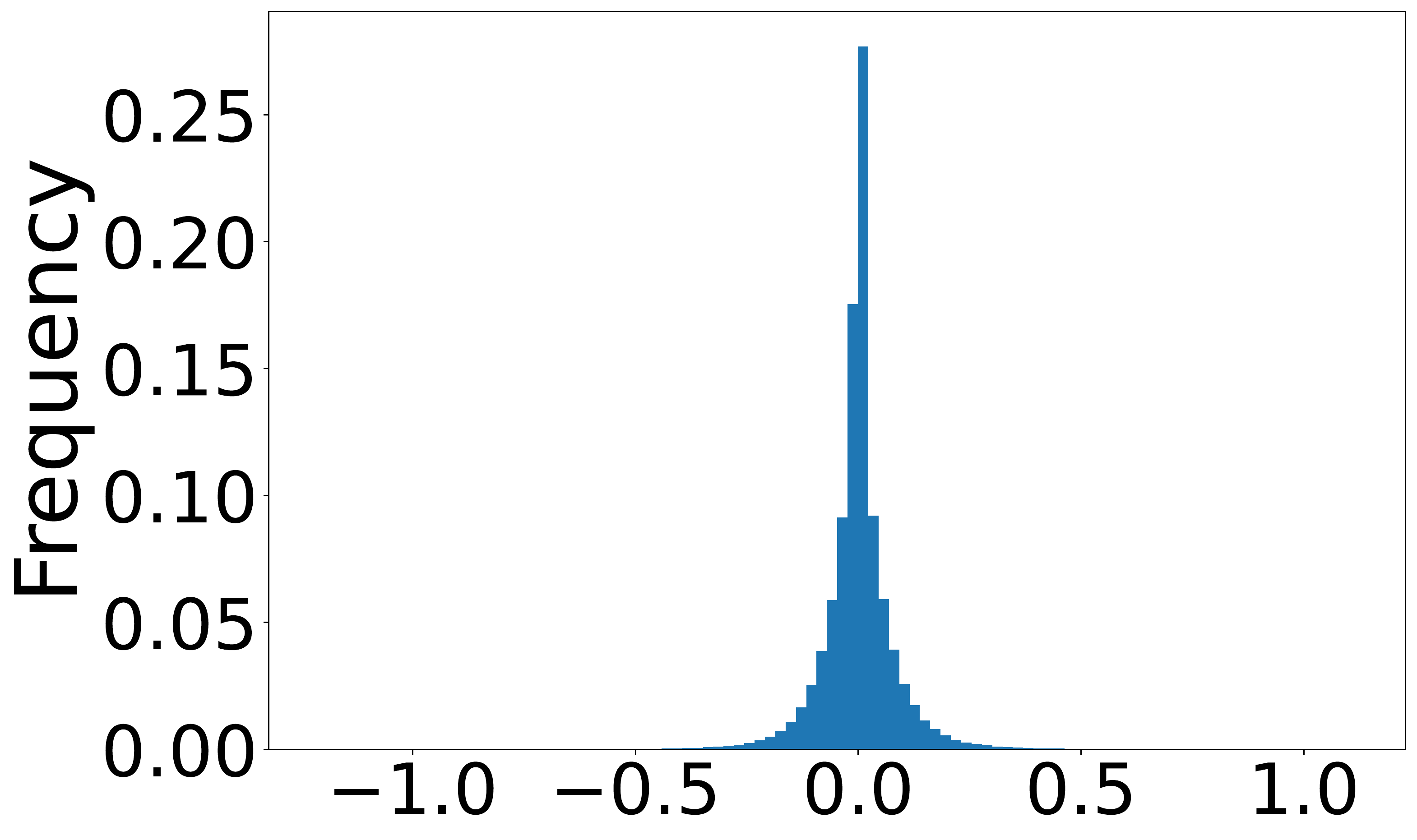}
    }
    \subfigure[Chebyshev]{
        \includegraphics[width=0.46\columnwidth]
        {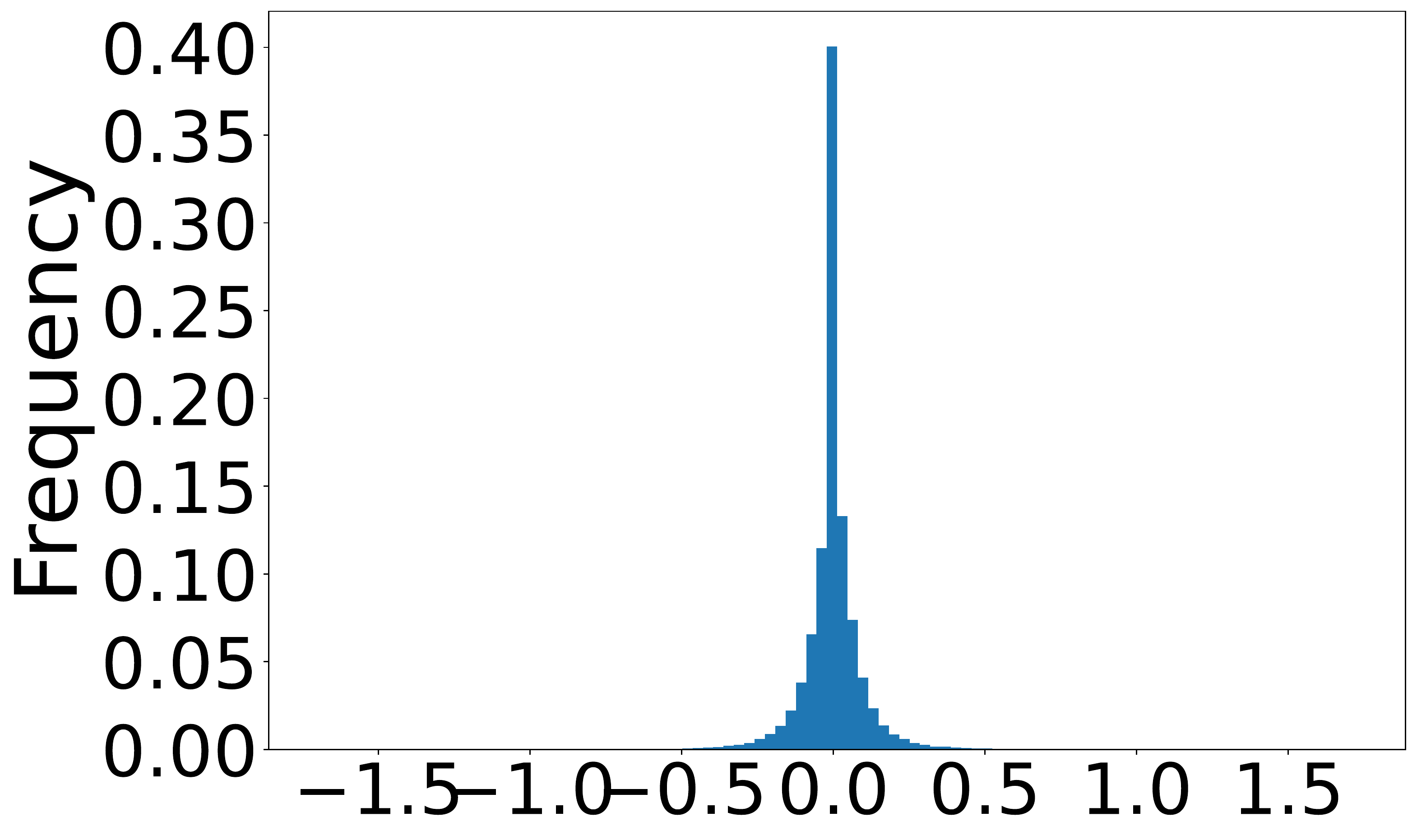}
    }
    \subfigure[SGC]{
        \includegraphics[width=0.46\columnwidth]
        {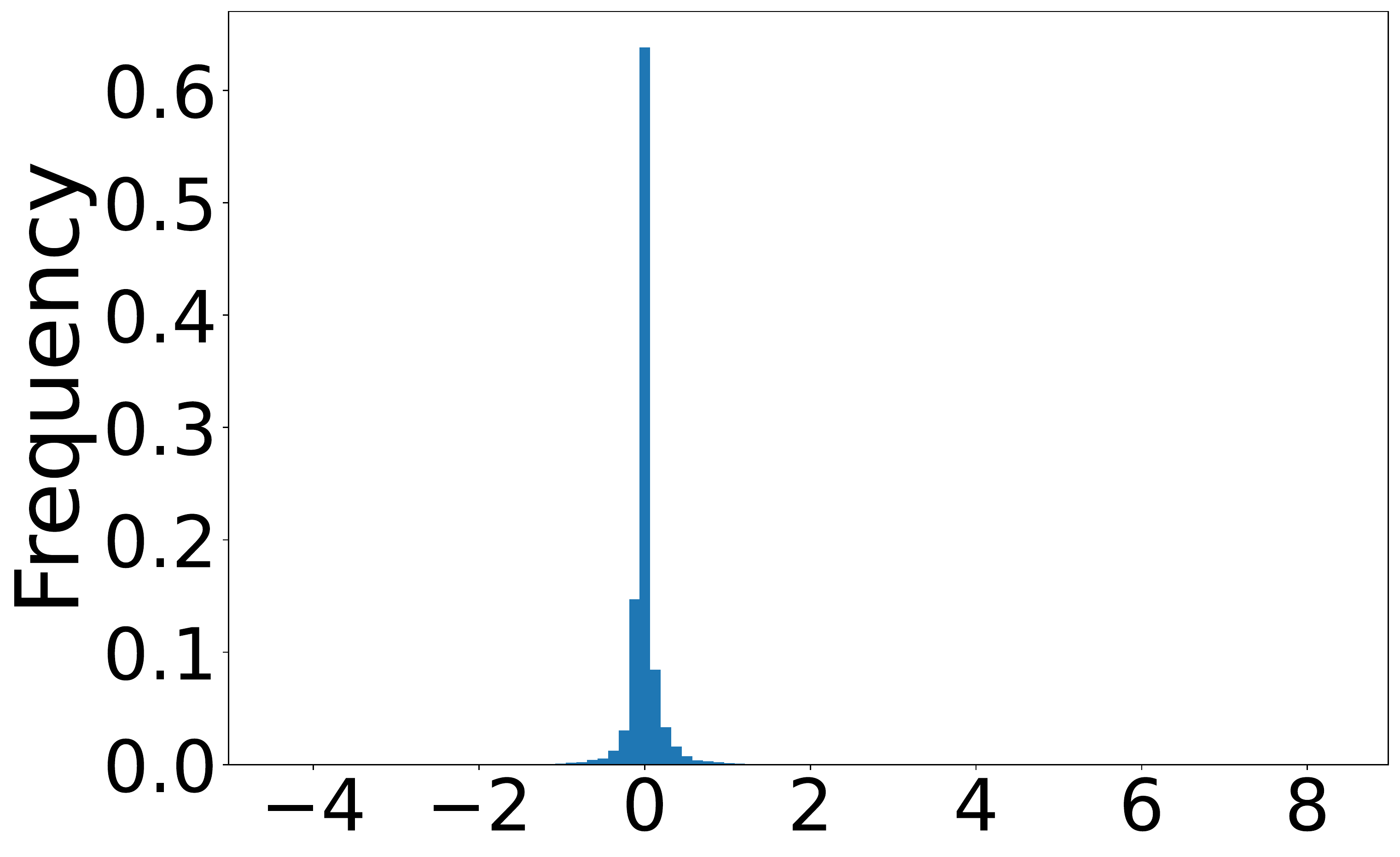}
    }
    \caption{Weight distribution of different GNNs models.}
    \label{fig:dist}
\end{figure}

We explore the model weight and activation clipping to mitigate the bit-flip errors in GNNs models. For the clipping threshold determination, however, different components can have dissimilar parameter distributions and it is hard to find a unified threshold for both weight and activation across different GNNs models, which have values of parameters in different ranges. For instance, we first analyze the distribution of parameters in the weights of different GNNs models, as Fig.~\ref{fig:dist} illustrated, which have variant ranges. On the other hand, the activation function utilized in GNNs models is $ReLU$, $ELU$, and $LogSoftmax$ functions, which introduces non-linear features in GNNs models and typically output values in the range of $[0, x]$, $(-1, x]$ and $(-\inf, 0)$ respectively. Consequently, without the loss of generality, we propose to first perform offline profiling on error-free GNNs models to measure the range of weight and activation values, including maximum and minimum values. Then, during the weight and activation clipping process, all the parameter values (in weights or activations) are clipped and limited to the range of the parameters of an error-free GNNs model.

% Although it may be out of the scope of this paper, a precise weight and activation clamping strategy will be considered in future work.

\subsection{GNN Topology-aware Activation Filtering}
\label{sec:filtering}
\begin{figure}[htbp!]
 	\centering
 	\includegraphics[width=0.45\textwidth]{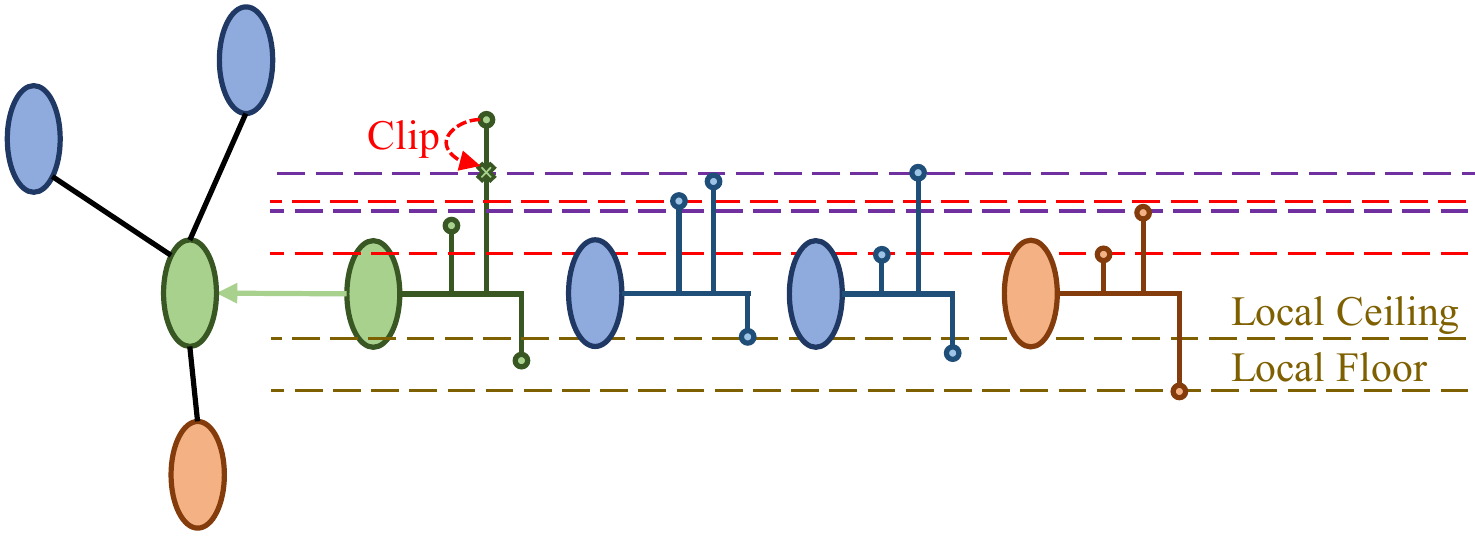}
 	\caption{GNN Topology-aware Activation Filtering on the target node (green) based on the feature value in its neighbors, where different dashed lines indicate different features. Each node (ellipse) has 3 different features (circles), and each feature will be filtered locally according to the corresponding feature values in the neighbor nodes.}
\label{fig:filter}
 \end{figure}

We propose an algorithm-level approach to mitigate the bit-flip errors in the activation outputs. In the GNNs model, the GNNs layers project the input feature into the representation space, while the representation maintains the topological information in the activation outputs, where the activation outputs have a graph structure. We propose an activation filtering approach that utilizes the topological structure of graphs as the guidance to diversely process the representation of different nodes. The proposed algorithm is indicated in Alg.~\ref{alg:filter}, where we use the neighbors of each node to smooth the corresponding node representation and to filter out the local errors. According to line 2 in our proposed algorithm, we select the nodes with a number of neighbors $N(\cdot)$ more than 1 that provides enough topological information for us to filter the value of the related node representations. Then from line 3 to line 8, we deploy an element-wise threshold computation, which means for each node in the activation graph, we find the maximum and minimum values of each element in the node representation, among all the neighbors of this node, as the ceiling and floor to clip the element in the node representation. For instance, if node $x$ has node representation $h(x) = [3, 6, -2]$, and node $x$ has neighbors $a, b \& c$ where $h(a) = [4, 4, -1]$, $h(b) = [2, 4, -2]$ and $h(c) = [2, 3, -3]$. An illustration is shown in Fig.~\ref{fig:filter}, where the representation of node $x$ will be clipped to $h'(x) = [3, 4, -2]$ according to the element-wise filtering algorithm.

\begin{algorithm}[htbp!]
\small
    \caption{Topology-aware Activation Filtering}
    \algrenewcommand\algorithmicrequire{\textbf{Input}}
    \algrenewcommand\algorithmicensure{\textbf{Output}}
    \begin{algorithmic}[1]
    \Require Activation Graph $G$
    \Ensure Filtered Activation Graph $G'$
    % \Ensure Fault Injected Model $M'$
    \State $G'$ $\leftarrow$ copy($G$)
    \For{Node $v$ in $G'$} \Comment{node level feature filtering}
        \If{$N(v) > 1$}
            \State $V$ is the hidden state representation of node $v$
            \State $h$ is the hidden state dimension of the GNNs model
            \For{Index $i$ in $[0, h - 1]$}
                \State Local Ceiling $C_l[i] \leftarrow max(N(v), axis = 0)$
                \State Local Floor $F_l[i] \leftarrow min(N(v), axis = 0)$
                \State Filtering $V'[i] \leftarrow Clip(V[i], C_l[i], F_l[i])$
            \EndFor
        \EndIf
    \EndFor \\
    \Return $G'$ 
    % \Comment{$M'$ is the fault injected model}
    % \State $M'(Node, Edge)$ \Comment{Activate all the registered hooks}
    \end{algorithmic}
    \label{alg:filter}
\end{algorithm}

% The comprehensive experimental results will be presented in Sec.\ref{sec:exp}, 
% In particular, for the software-dominated mitigation approaches, like clipping and filtering, we first mitigate NaN values to ``0'' since the NaN value is not a valid operation number within the clipping and filtering calculations.

% Combined with Razor circuitry, both methods can be implemented with 0.3\% silicon area and 12.8\% power overheads on a single-port SRAM~\cite{reagen2016minerva}. The 
\section{Experimental Results}
\label{sec:exp}
\subsection{Experimental Setup}
%An overview of GNNs models and graph information employed in PyGFI shows in Table.\ref{tab:gnn}.
We use three widely-used citation network datasets as the benchmarks for the GNNs models, including Cora, Citeseer, and Pubmed~\cite{kipf2016semi}. The nodes in the datasets represent document features and the edges represent the citation links. Since the node classification task follows a semi-supervised learning nature, the label rate of each dataset is $0.052$, $0.036$ and $0.003$ respectively, which means only 3\%-5\% are labeled for the purpose of training in the three datasets/. Table~\ref{tab:graph} shows the dataset statistics including the number of nodes and edges for each dataset.

We use four popular GNNs models: GCN~\cite{kipf2016semi}, GAT~\cite{velivckovic2017graph}, Chebyshev~\cite{defferrard2016convolutional}, and SGC~\cite{wu2019simplifying}. Those GNNs models share a similar architecture, manifested as a 2-layer GNN structure illustrated in Fig.~\ref{fig:pygfi}, but have different information aggregation approaches. To be specific, GCN model, named graph convolutional network, deploys message passing through a normalized adjacency matrix to aggregate topological information from neighbors, while SGC is a simplified version of GCN model that removes activation layers and compresses two GCN layers into one to speed up the GCNs computation. On the other hand, Chebyshev utilizes localized spectral filtering and achieves the graph convolutional process through approximated Chebyshev polynomials. Moreover, the GAT model, named graph attention network, introduces the attention mechanism into GNNs computation and employs concatenation operation to combine topological information.

\begin{table}[htbp!]
\caption{Graph Dataset Statistics}
  \centering
    \begin{tabular}{cccc}
    \toprule
    % \multirow{6}*{\rotatebox{90}{GNNs Model Information}}
    Dataset     & Cora   & Citeseer & Pubmed\\
    \midrule
    \#Nodes      &  2708  &  3327  &  19717 \\
    \#Edges      &  5429  &  4732  &  44338 \\
    \#Features   &  1433  &  3703  &  500   \\
    \#Categories & 7      &  6     &  3     \\
    \bottomrule  
    \end{tabular}
  \label{tab:graph}
  %\vspace{-0.5cm}
\end{table}

\begin{table}[htbp!]
% \small
\caption{Accuracy \& model size of different GNNs models}
  \centering
  % \hspace{-20pt}
    \begin{tabular}{cccc}
    \toprule
    % \multirow{6}*{\rotatebox{90}{GNNs Model Information}}
     {GNNs model}    & Total Weights & Total Activations & Accuracy\\
     \midrule
     {GCN (Cora)}                 & 23K  & 62K   &0.814\\
     {GCN (Citeseer)}             & 59K  & 73K   &0.707\\ 
     {GCN (Pubmed)}               & 8K   & 375K  &0.793\\ 
     {GAT (Cora)}                 & 92K  & 192K  &0.827\\ 
     {GAT (Citeseer)}             & 237K & 232K  &0.718\\ 
     {GAT (Pubmed)}               & 33K  & 1321K &0.786\\ 
     {SGC (Cora)}                 & 10K  & 19K   &0.786\\ 
     {SGC (Citeseer)}             & 22K  & 20K   &0.73 \\ 
     {SGC (Pubmed)}               & 1.5K & 59K   &0.777\\ 
     {Chebyshev (Cora)}           & 46K  & 62K   &0.806\\ 
     {Chebyshev (Citeseer)}       & 119K & 73K   &0.694\\ 
     {Chebyshev (Pubmed)}         & 16K  & 375K  &0.767\\ 
    \bottomrule  
    \end{tabular}
  \label{tab:gnn}
  %\vspace{-0.5cm}
\end{table}

\begin{table}[htbp!]
% \small
\caption{Weight \& activation value range of GNNs models}
  \centering
  % \hspace{-20pt}
    \begin{tabular}{cccc}
    \toprule
    % \multirow{6}*{\rotatebox{90}{GNNs Model Information}}
     {GNNs model}                & Component & Minimum Value & Maximum Value\\
     \midrule
     {GCN (Cora)}               & Weights     &-1.713      &1.82\\
     {GCN (Citeseer)}           & Weights     &-1.83       &1.901\\ 
     {GCN (Pubmed)}             & Weights     &-1.337      &1.383\\ 
     {GAT (Cora)}               & Weights     &-0.892      &0.911\\ 
     {GAT (Citeseer)}           & Weights     &-0.925      &0.868\\ 
     {GAT (Pubmed)}             & Weights     &-0.807      &0.835\\ 
     {SGC (Cora)}               & Weights     &-4.005      &6.346\\ 
     {SGC (Citeseer)}           & Weights     &-1.666      &6.358\\ 
     {SGC (Pubmed)}             & Weights     &-4.388      &8.339\\ 
     {Chebyshev (Cora)}         & Weights     &-1.266      &1.245\\ 
     {Chebyshev (Citeseer)}     & Weights     &-1.391      &1.303\\ 
     {Chebyshev (Pubmed)}       & Weights     &-1.285      &1.26\\ 

     {GCN (Cora)}               & Activations &-24.601     &1.513\\
     {GCN (Citeseer)}           & Activations &-13.158     &0.746\\ 
     {GCN (Pubmed)}             & Activations &-11.422     &1.042\\ 
     {GAT (Cora)}               & Activations &-5.671      &0.237\\ 
     {GAT (Citeseer)}           & Activations &-2.348      &0.037\\ 
     {GAT (Pubmed)}             & Activations &-3.98       &0.14\\ 
     {SGC (Cora)}               & Activations &-3.648      &-0.288\\ 
     {SGC (Citeseer)}           & Activations &-2.160      &-1.049\\ 
     {SGC (Pubmed)}             & Activations &-3.548      & -0.076\\ 
     {Chebyshev (Cora)}         & Activations &-33.131     &1.348\\ 
     {Chebyshev (Citeseer)}     & Activations &-15.259     &0.621\\ 
     {Chebyshev (Pubmed)}       & Activations &-19.381     &0.894\\
    \bottomrule  
    \end{tabular}
  \label{tab:dist}
  %\vspace{-0.5cm}
\end{table}

Table~\ref{tab:gnn} shows the statistics of each GNN-Dataset combination, including information such as the original accuracy and number of weights and activations. Note that due to the very nature of GNN, the number of parameters will vary based on different datasets. For example, the (average) number of edges per node in variant datasets is not identical, so the number of weights/activation outputs for the GNNs are also different, which will potentially affect the resiliency of the specific GNN-dataset combination. Furthermore, we offline profile the GNNs models and measure the range of weight and activation values, as Table.~\ref{tab:dist} den oted. All the datasets and the GNNs models we utilized and implemented are based on PyTorch 2.0 and PyG library~\cite{fey2019fast}.

For each GNN-dataset combination, we sweep 8-bit error rates equally spaced on a logarithmic scale from $10^{-9}$ to $10^{-2}$, e.g., $10^{-9}$, $10^{-8}$, etc. At each error rate, we repeat our experiment 10 times and report the mean accuracy. To assess the sensitivity of different components of GNN, we deploy error injections at different parts of GNNs models:
(1) \textit{Model-wise}: we inject errors to all the weights in the model. (2) \textit{GNN-1}: we inject errors only to weights in the first layer. Note that the layer name is GNN-specific; so if the GNNs model is GCN/GAT, then the layer name is GCN-1/GAT-1. (3) \textit{GNN-2}: We inject errors only to weights in the second layer; same as GNN-1, the name of this layer is also network-specific, e.g., GCN-2, GAT-2, etc. Also note that unlike many popular CNNs, many widely-used GNNs have no more than 2 layers, e.g., GCN only has two GCN convolution layers~\cite{kipf2016semi}, i.e., GCN-1 and GCN-2. (4) Additionally, we also assess the sensitivity of the activation outputs by injecting errors only to them, referred to as \textit{Activation}.

%  \begin{figure}[htbp!]
%  	\centering
%      % \hspace{-20pt}
%  	\includegraphics[width=0.5\textwidth]{./figures/param_dist.pdf}
%  	\caption{Weight \& activation value distribution of different GNNs}
% \label{fig:dist}
%  \end{figure}

%For example, we inject errors to the enable up to 6 different fault configurations: model-wise injection, layer-1 injection, layer-2 injection, and injection to 

\begin{figure*}[htbp!]
    \centering

    \subfigure[GCN - Cora]{
        \includegraphics[width=0.6\columnwidth]{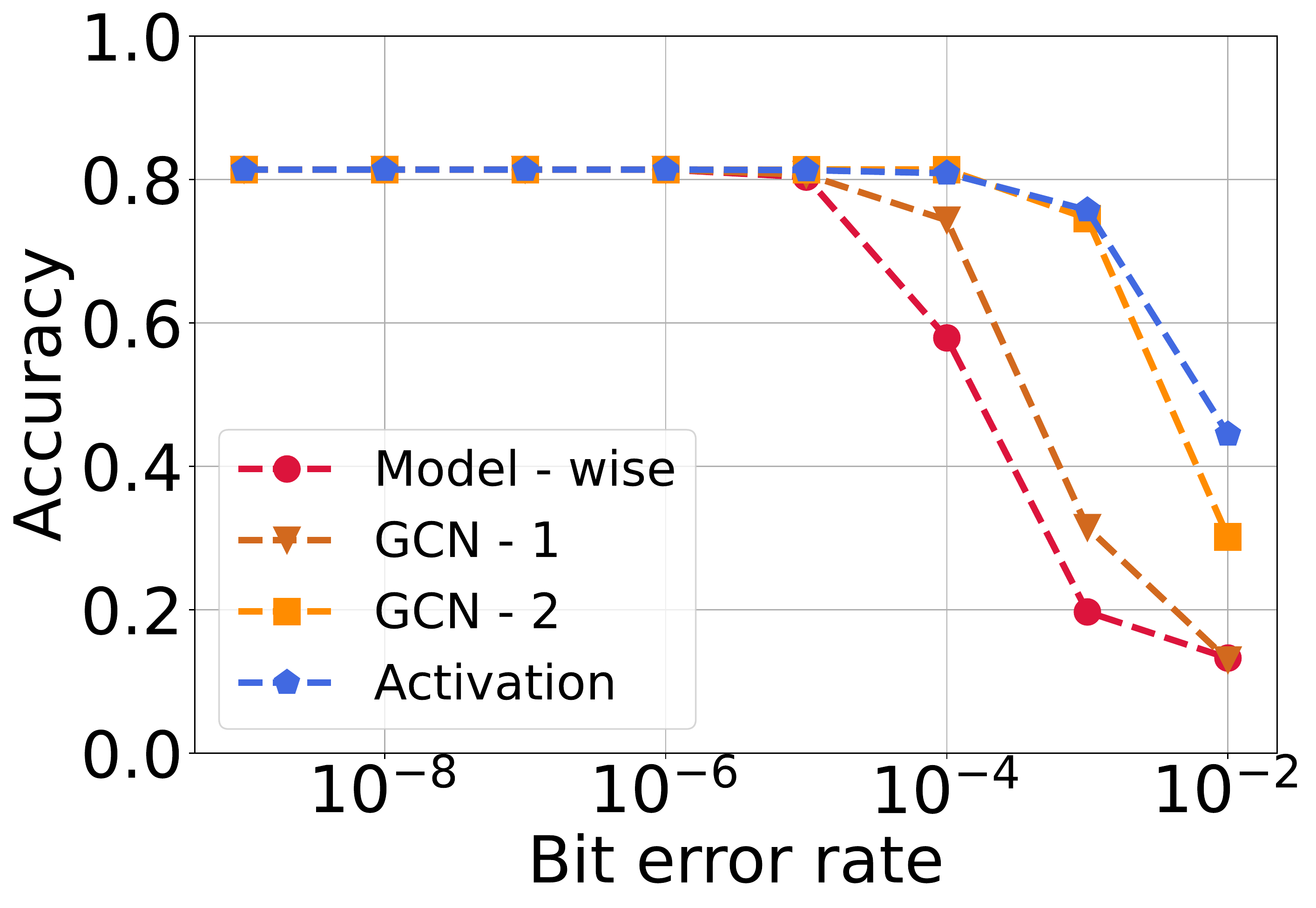}
    }
    \hskip -1em 
    \subfigure[GCN - Citeseer]{
        \includegraphics[width=0.6\columnwidth]{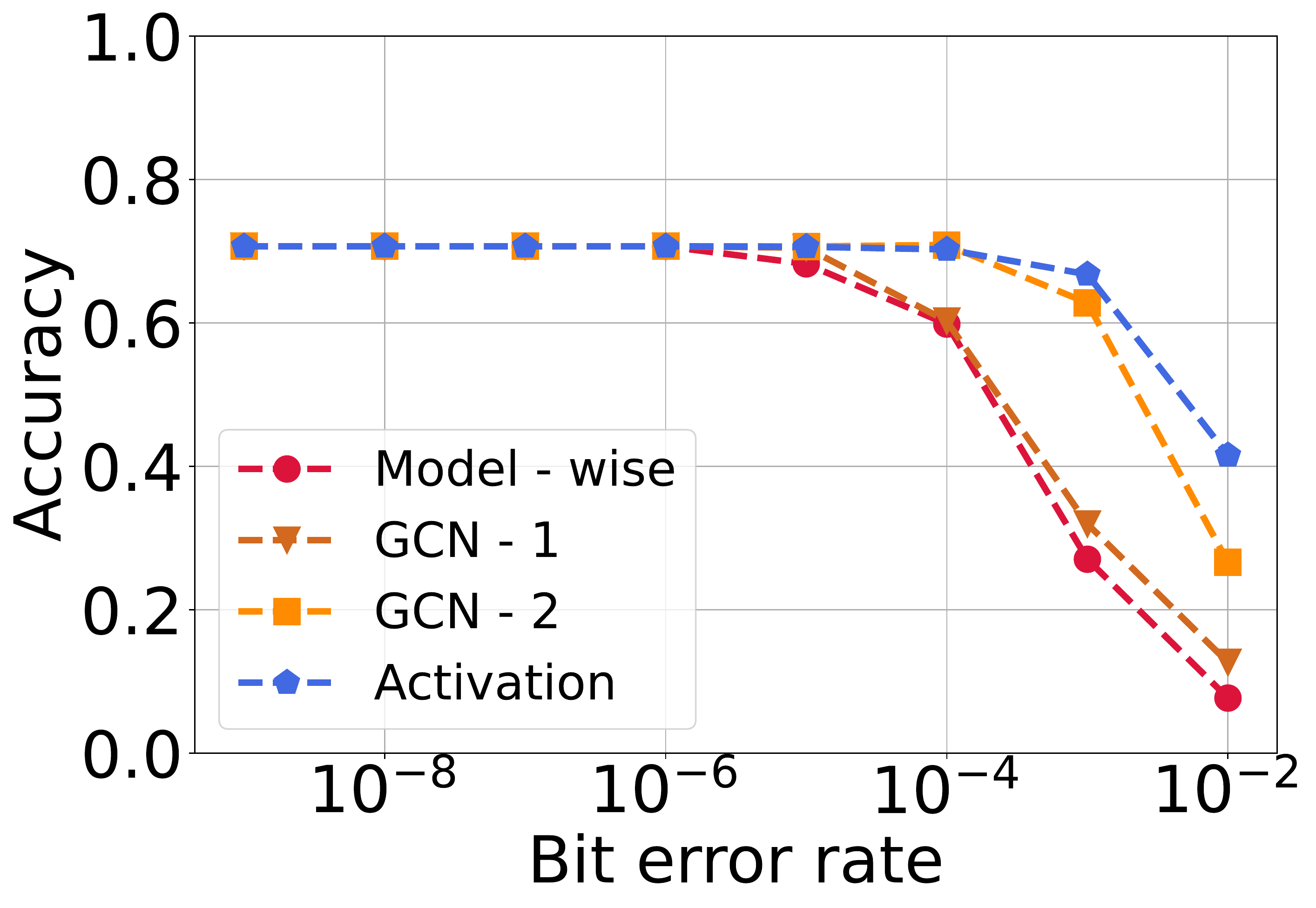}
    }
    \hskip -1em 
    \subfigure[GCN - Pubmed]{
        \includegraphics[width=0.6\columnwidth]{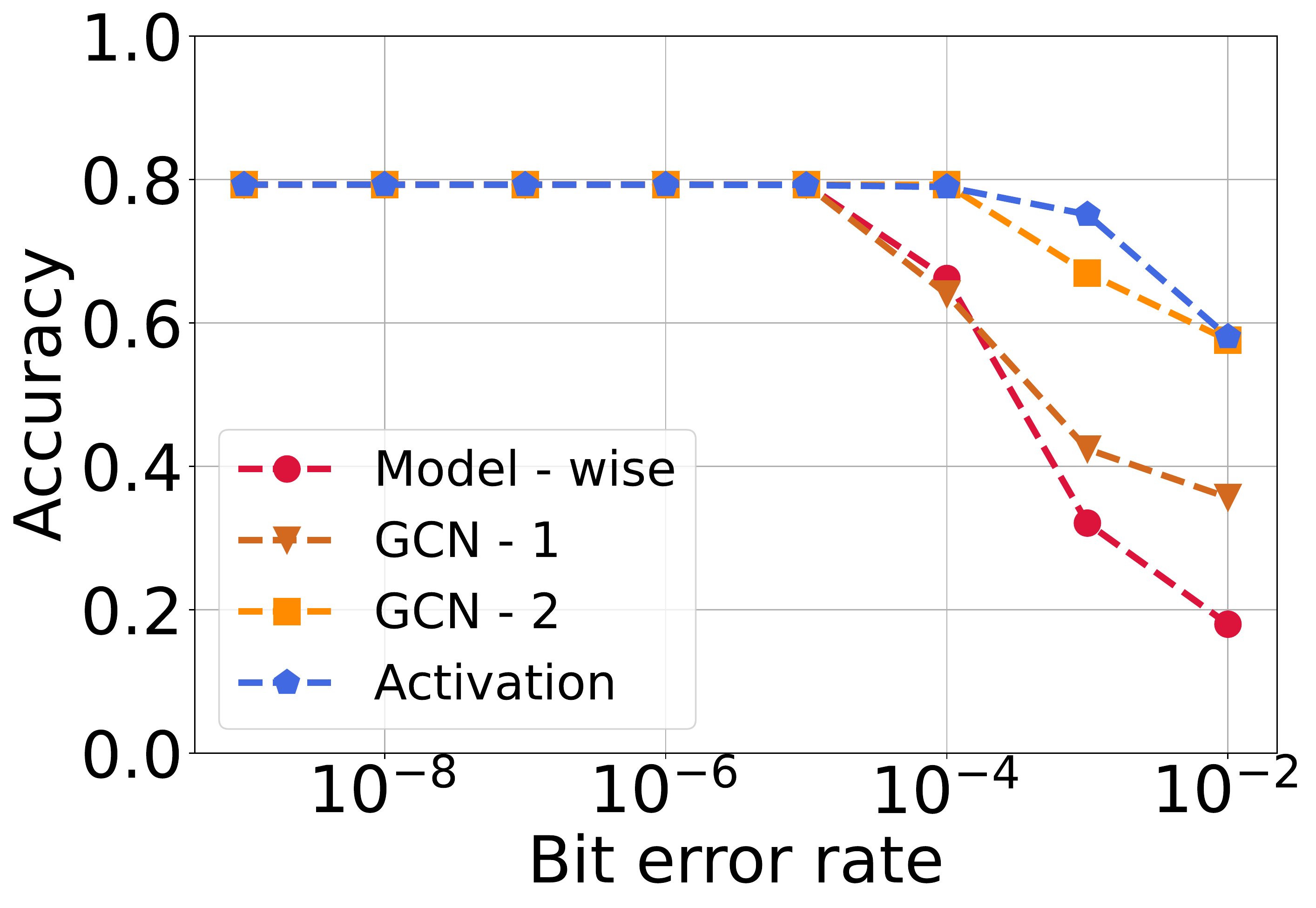}
    }
    \hskip -1em     
    \subfigure[GAT - Cora]{
        \includegraphics[width=0.6\columnwidth]{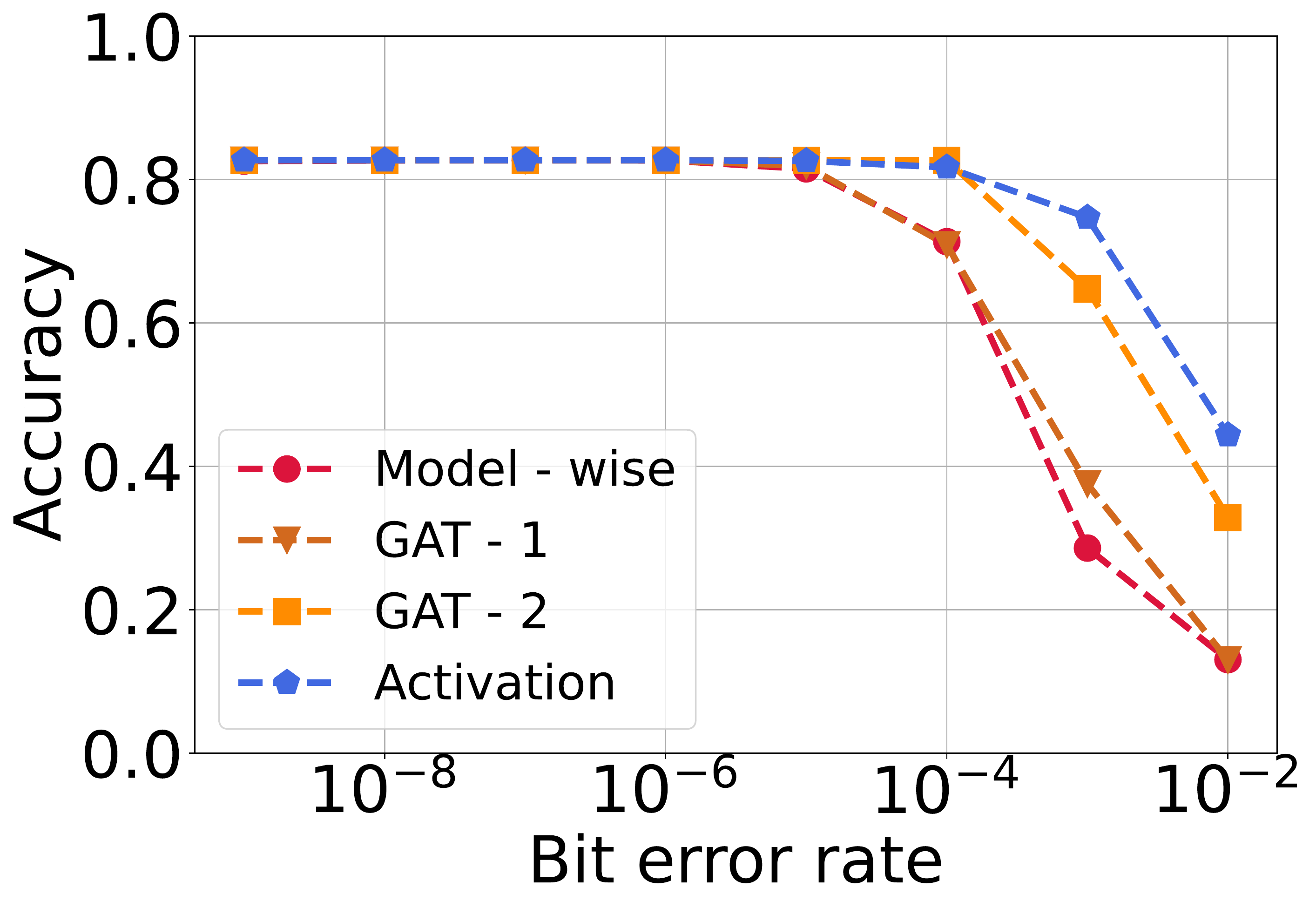}
    }
    \hskip -1em 
    \subfigure[GAT - Citeseer]{
        \includegraphics[width=0.6\columnwidth]{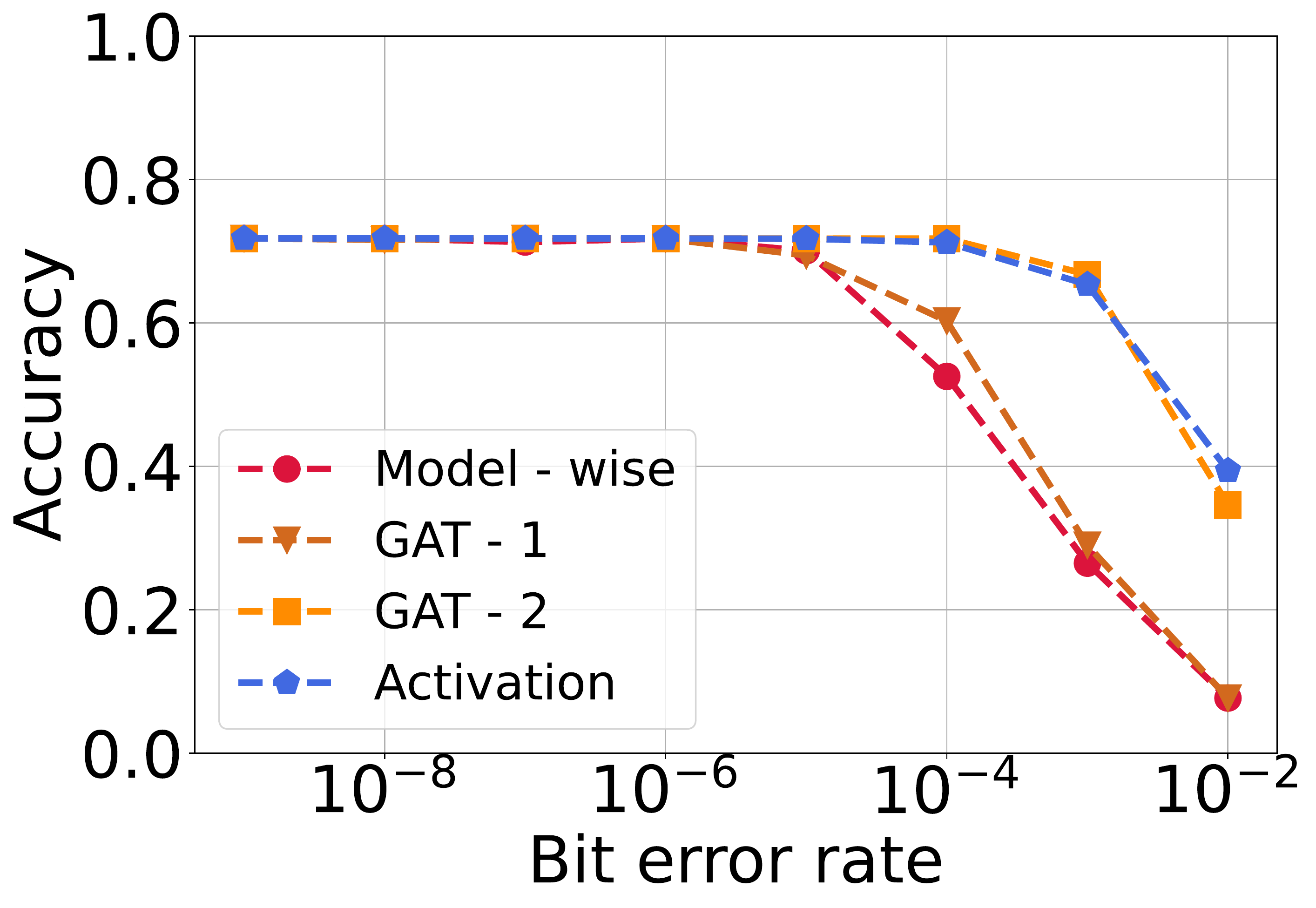}
    }
    \hskip -1em 
    \subfigure[GAT - Pubmed]{
        \includegraphics[width=0.6\columnwidth]{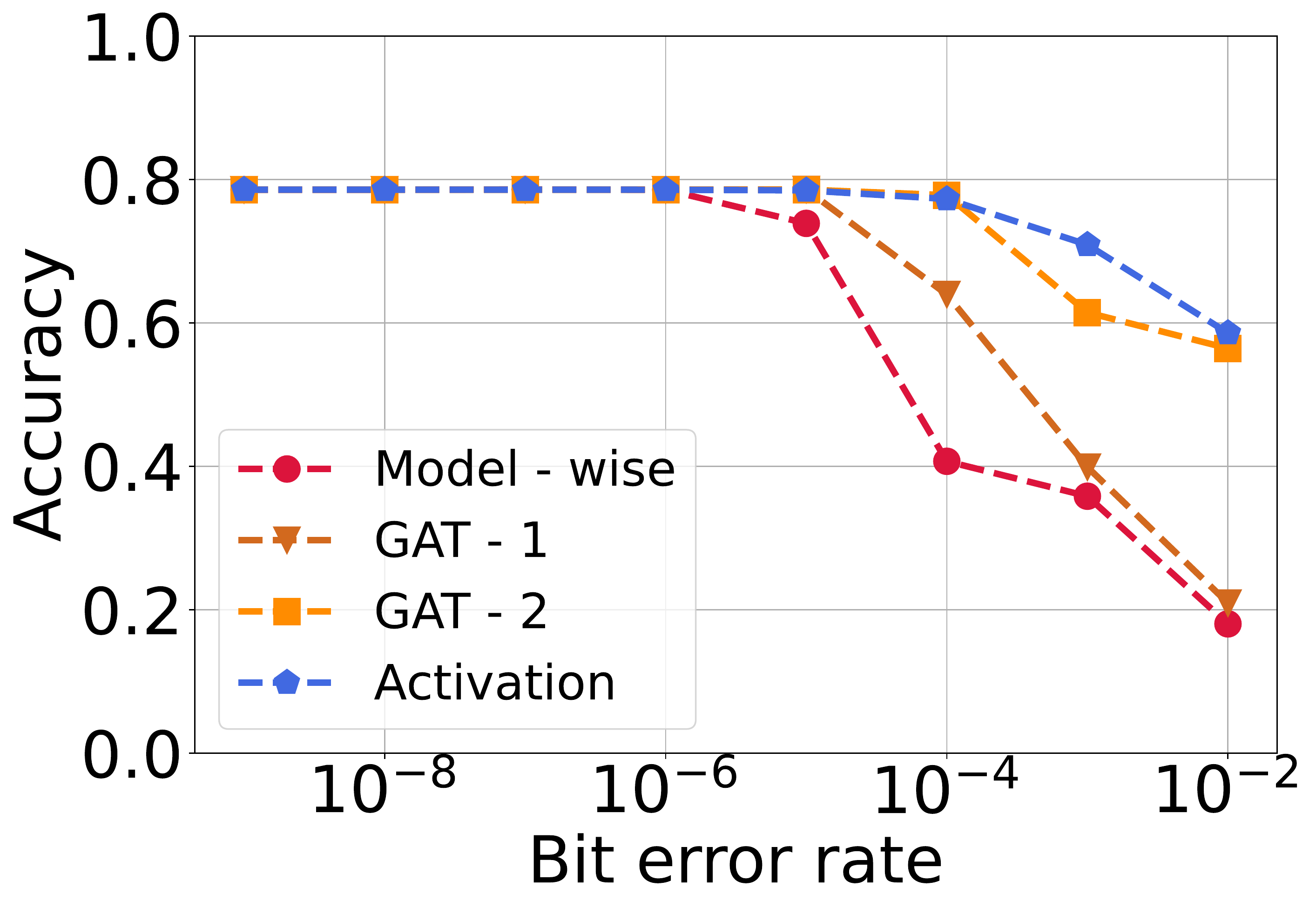}
    }
    \subfigure[Chebyshev - Cora]{
        \includegraphics[width=0.6\columnwidth]{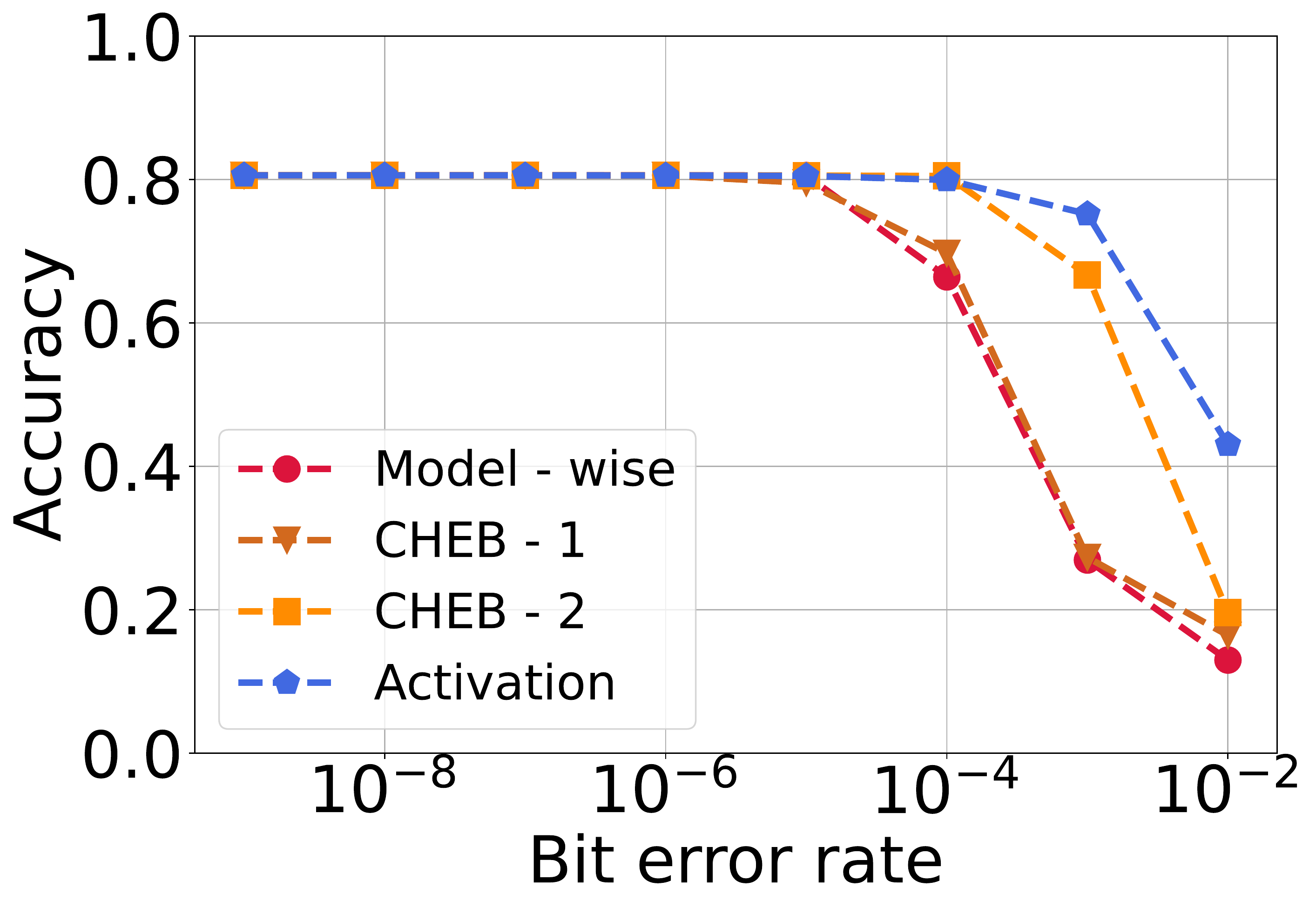}
    }
    \hskip -1em 
    \subfigure[Chebyshev - Citeseer]{
        \includegraphics[width=0.6\columnwidth]{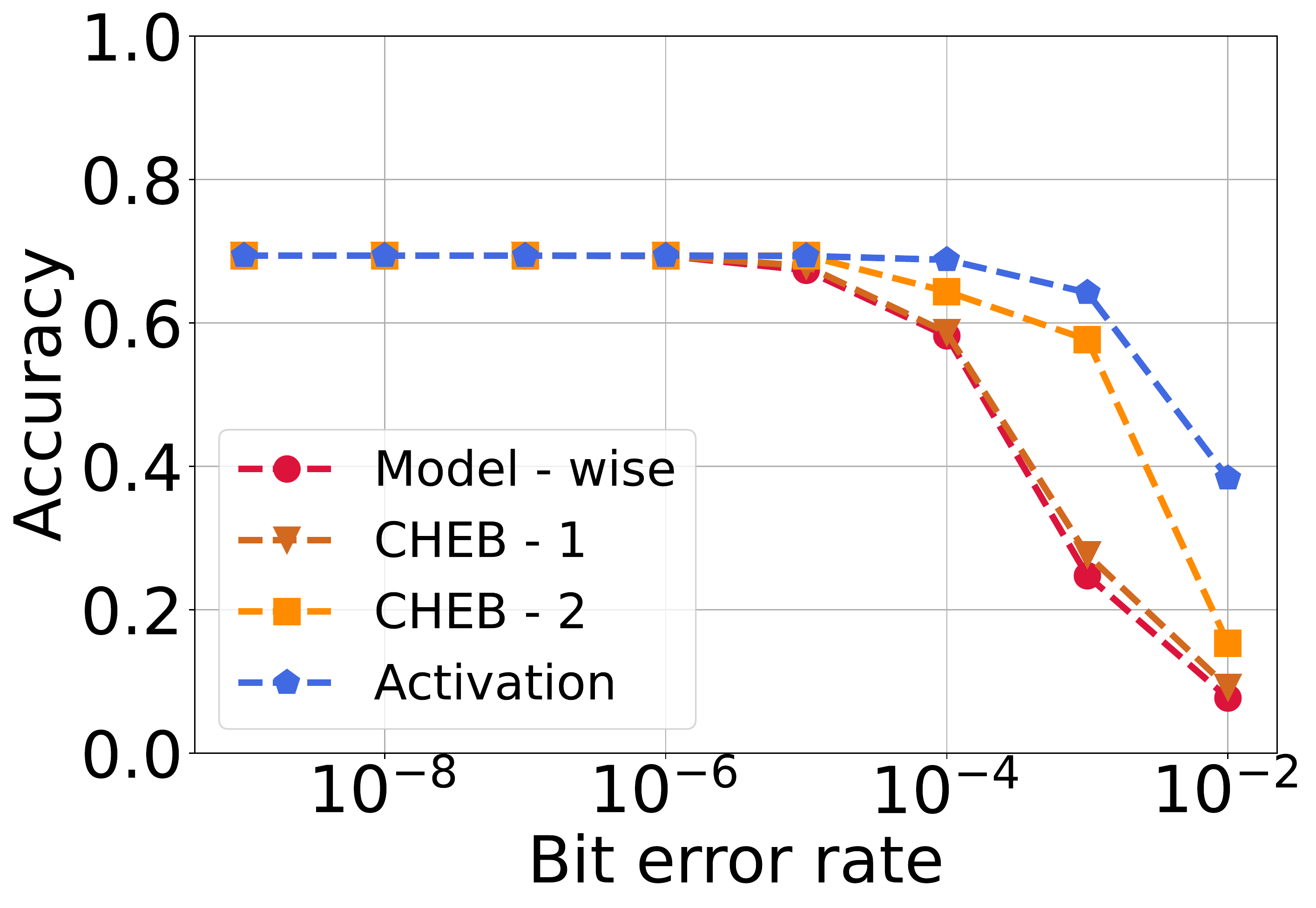}
    }
    \hskip -1em 
    \subfigure[Chebyshev - Pubmed]{
        \includegraphics[width=0.6\columnwidth]{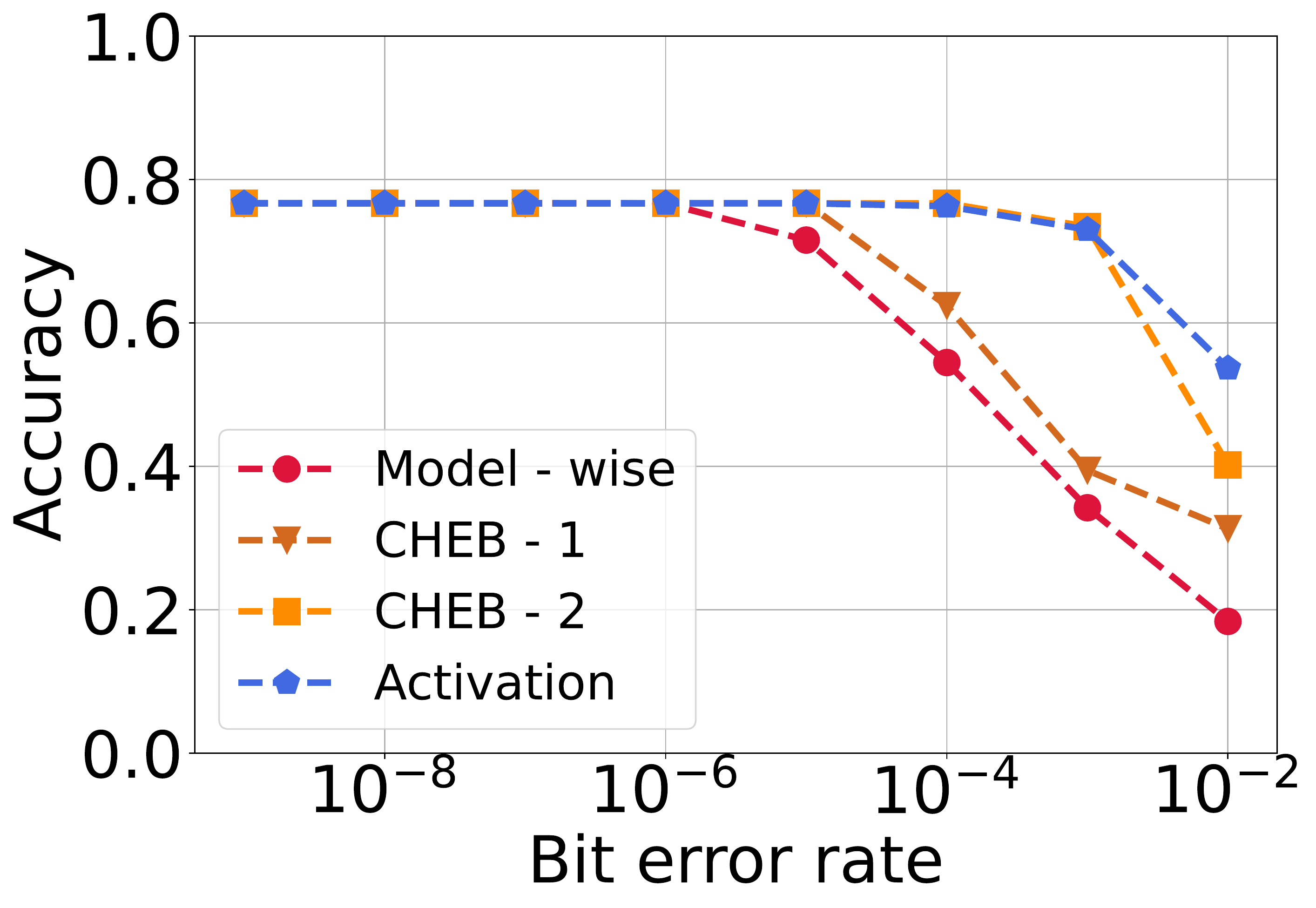}
    }
    \hskip -1em    
    \subfigure[SGC - Cora]{
        \includegraphics[width=0.6\columnwidth]{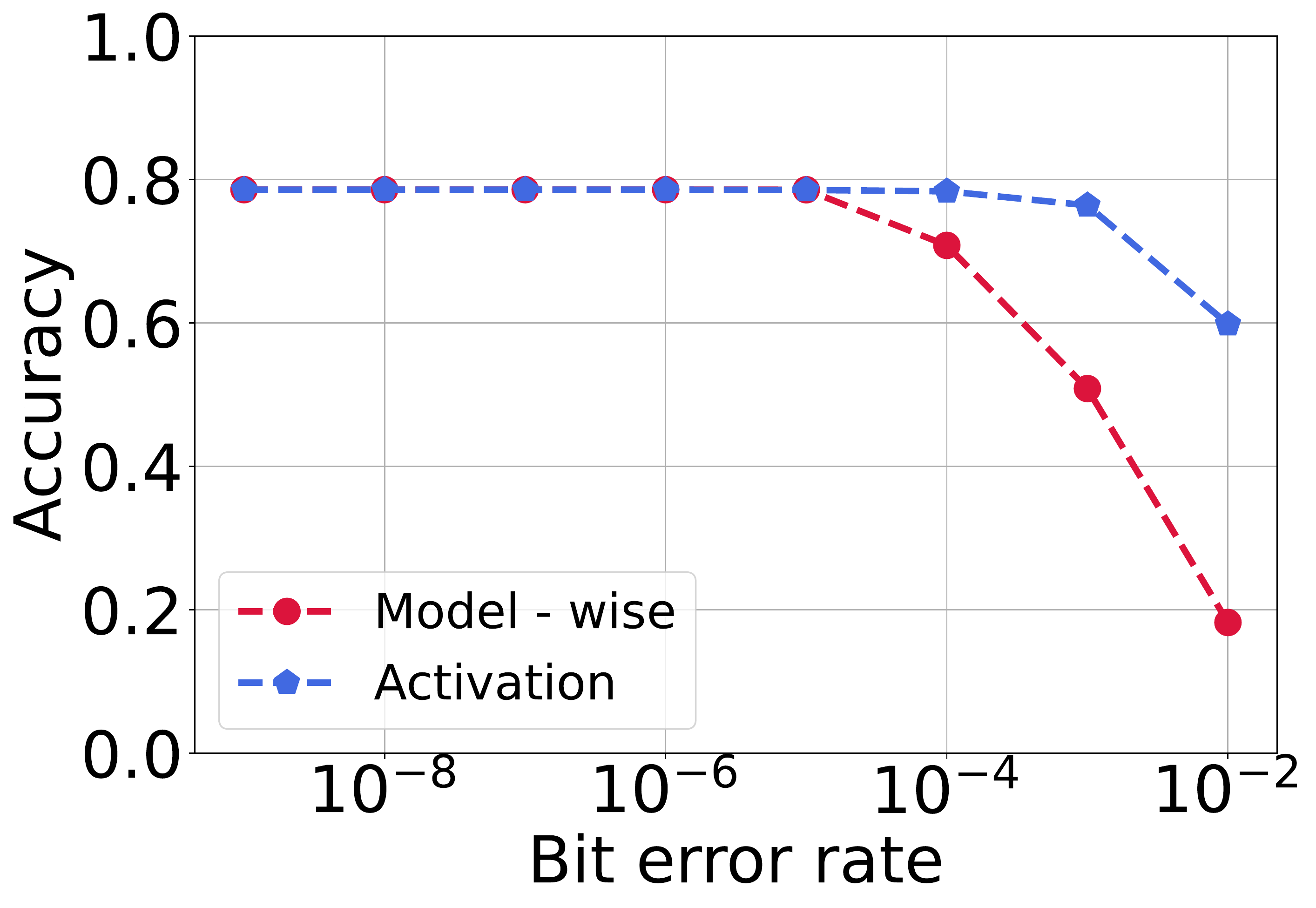}
    }
    \hskip -1em 
    \subfigure[SGC - Citeseer]{
        \includegraphics[width=0.6\columnwidth]{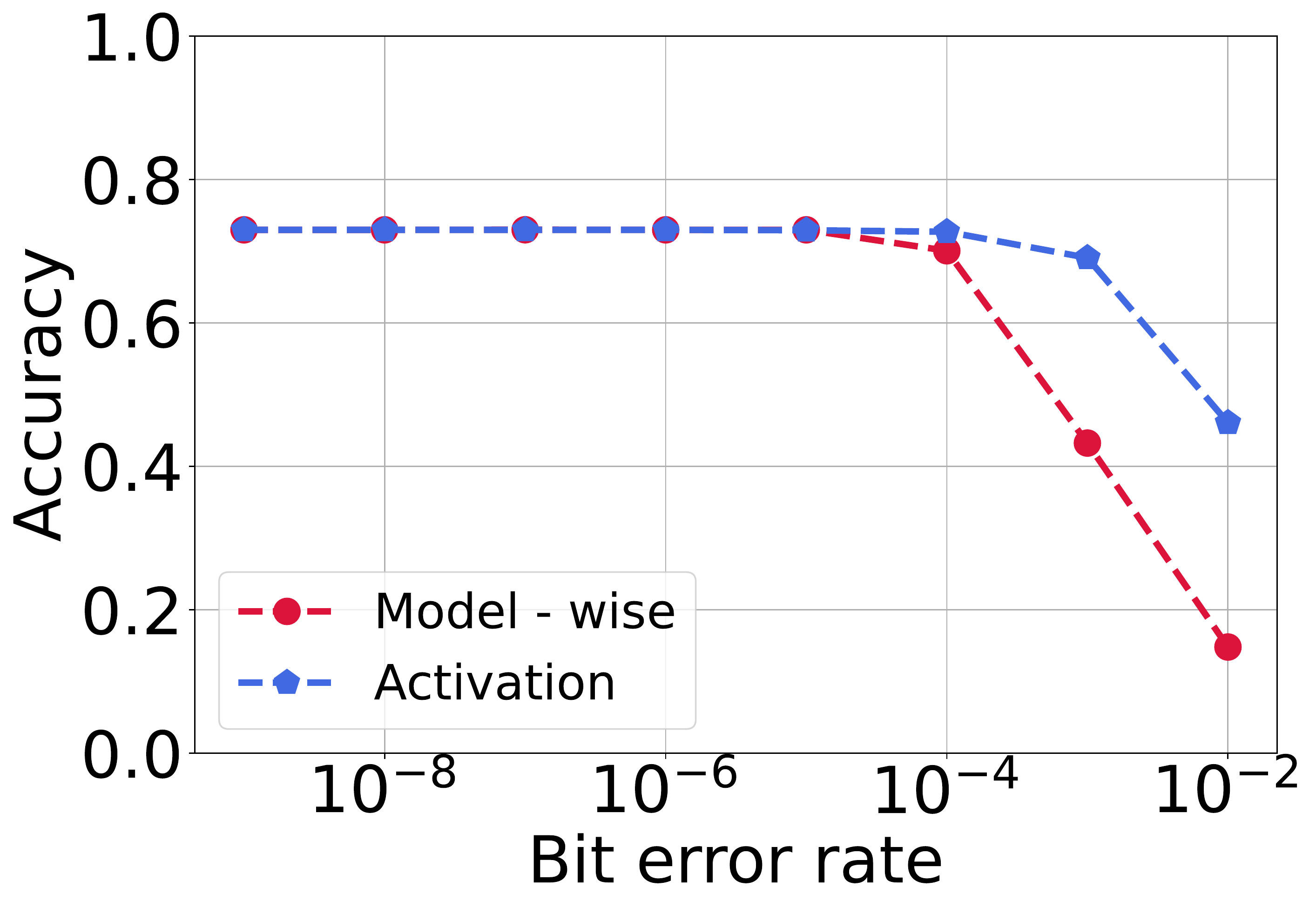}
    }
    \hskip -1em 
    \subfigure[SGC - Pubmed]{
        \includegraphics[width=0.6\columnwidth]{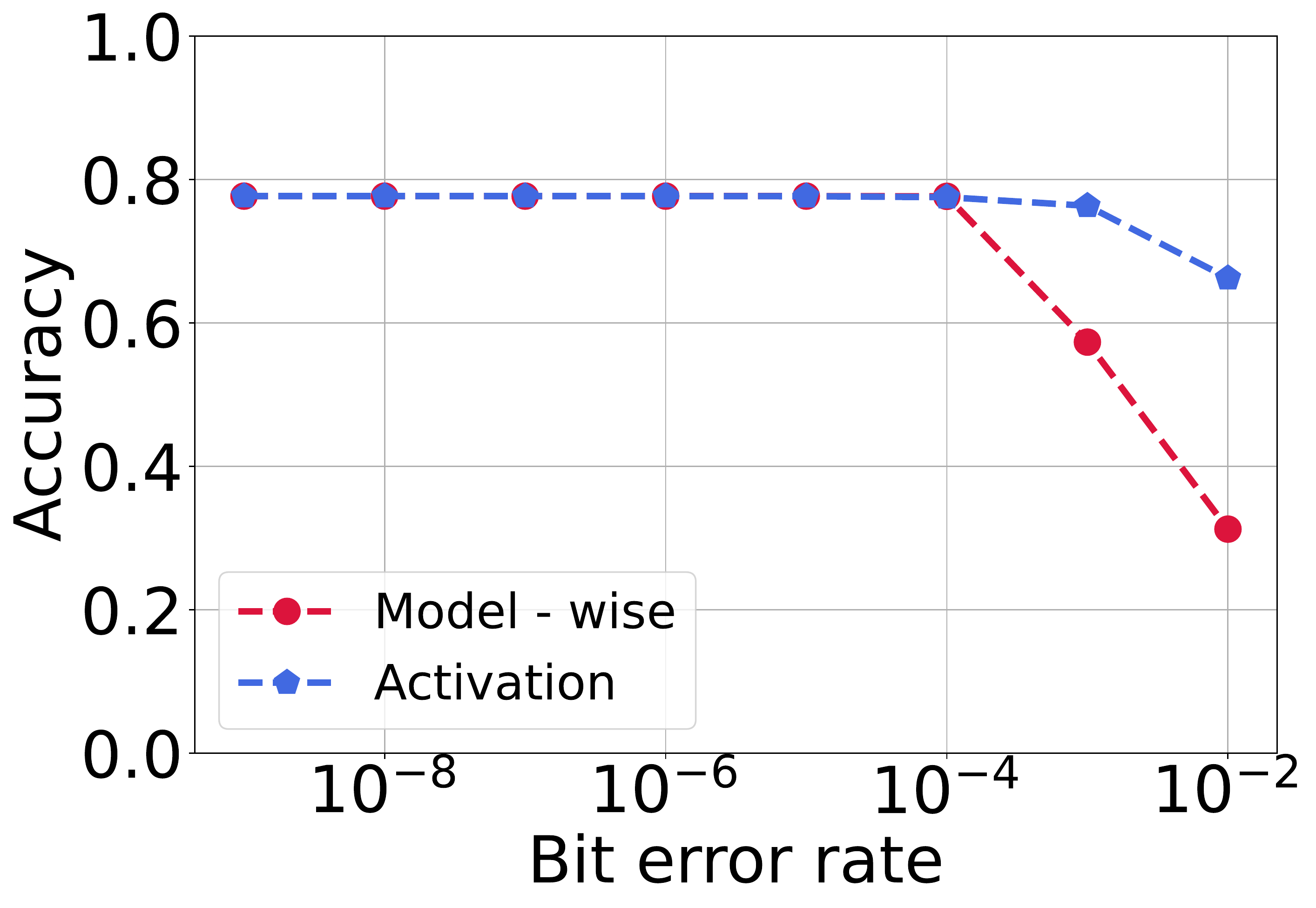}
    }
    %\vspace{-5pt}
    \caption{robustness of different GNNs models under BER in the range of $[10^{-9}, 10^{-2}]$.}
    %\vspace{-0.5cm}
    \label{fig:tsd_comp}
    %\vspace{-15pt}
\end{figure*}

\subsection{GNN Robustness to Hardware Errors}

Fig.~\ref{fig:tsd_comp} shows the robustness results of GNNs across our evaluation space. In this section, we focus on the model-wise curve because we focus on the robustness of GNN as an entirety first, and we will discuss the sensitivity of different components, e.g., GCN-1/2, in the next section. 
%Overall, we observe orders of magnitude variance of error robustness across HDC models and their configurations, which can be leveraged to explore the design optimization of HDC systems.

\textbf{RQ1:} Are GNNs models robust to hardware errors?

\textbf{Ans:} Based on Fig.~\ref{fig:tsd_comp}, we can see that GNNs are inherently robust to hardware errors to a considerable extent. For most GNNs, we do not see a noticeable accuracy drop until the fault rate hits $10^{-6}$, e.g., GCN-Cora (Fig.~\ref{fig:tsd_comp} (a)), and some even go up to $10^{-4}$, e.g., SGC-Pubmed (Fig.~\ref{fig:tsd_comp} (l)). (For convenience, these fault rates are named as ``cutoff fault rates''.) This can be regarded as strong robustness, especially considering that conventional computing typically requires negligible error rates (e.g., $< 10^{-15}$~\cite{jesd2182010solid}) for correct execution. This strong resiliency can be potentially leveraged to design efficient GNN systems. Nevertheless, after cutoff rates, we do see a sharp drop in accuracy as the fault rate increases. This suggests that designers need to be extremely careful in balancing the design decisions between the performance-efficiency-robustness tradeoffs, which highlights the value of GNN robustness research.

\textbf{RQ2:} Does the impact of hardware errors on GNN accuracy vary across different models and application datasets?

\textbf{Ans:} \textbf{Yes, on both}. We observe a significant variance in robustness across models. For \textbf{SGC} (Fig.~\ref{fig:tsd_comp} (j) - (l)), the accuracy drop is much slower than in other models. For example, in Fig.~\ref{fig:tsd_comp} (l), we do not see a notable accuracy drop for SGC-Pubmed even at $10^{-4}$, while for other models under the same dataset, e.g., GAT and Chebyshev, they have already started facing accuracy drop at $10^{-6}$. This is two orders of magnitude difference. Actually, across all the datasets, SGC exhibits notably higher robustness than all the other models: SGC has cutoff rates beyond $10^{-5}$, while other models have cutoff rates at $10^{-6}$ or even smaller. 
%fault rate when we start to see accuracy drop. This suggests that when designing resilient GNN systems, we need to consider specific GNNs models to make design choices. 

The robustness differences between models are reasonable because each GNNs model has a different architecture, number of parameters, etc. In particular, based on Table~\ref{tab:gnn}, we can observe that SGC has the fewest parameters, which makes the errors unlikely to happen even when the fault rate is relatively higher compared to other models. For example, under Pubmed, SGC only has $1.5K$ weights, which is significantly less than $8K, 33K, 16K$ for GCN, GAT, and Chebyshev.

We also observe a significant variance in robustness across application datasets. For \textbf{GCN} (Fig.~\ref{fig:tsd_comp} (a) - (c)), the accuracy drop of GCN-Pubmed is considerably slower than GCN-Cora/Citeseer. We do not see a notable accuracy drop for GCN-Pubmed until $10^{-5}$, while for GCN-Cora/Citeseer, they start to drop at $10^{-6}$ fault rate. The robustness difference between different datasets also comes from their direct impact on the GNN structures including the number of parameters. For example, based on Table~\ref{tab:gnn}, GCN-Pubmed has in total $8K$ weights while GCN-Cora/Citeseer have $23K$ and $59K$, respectively. %This suggests that when designing resilient GNN systems, we need to consider specific application datasets to make design choices. 

%\subsection{robustness Results of GNNs}

\begin{figure*}[htbp!]
    \centering

    \subfigure[GCN]{
        \includegraphics[width=0.473\columnwidth]{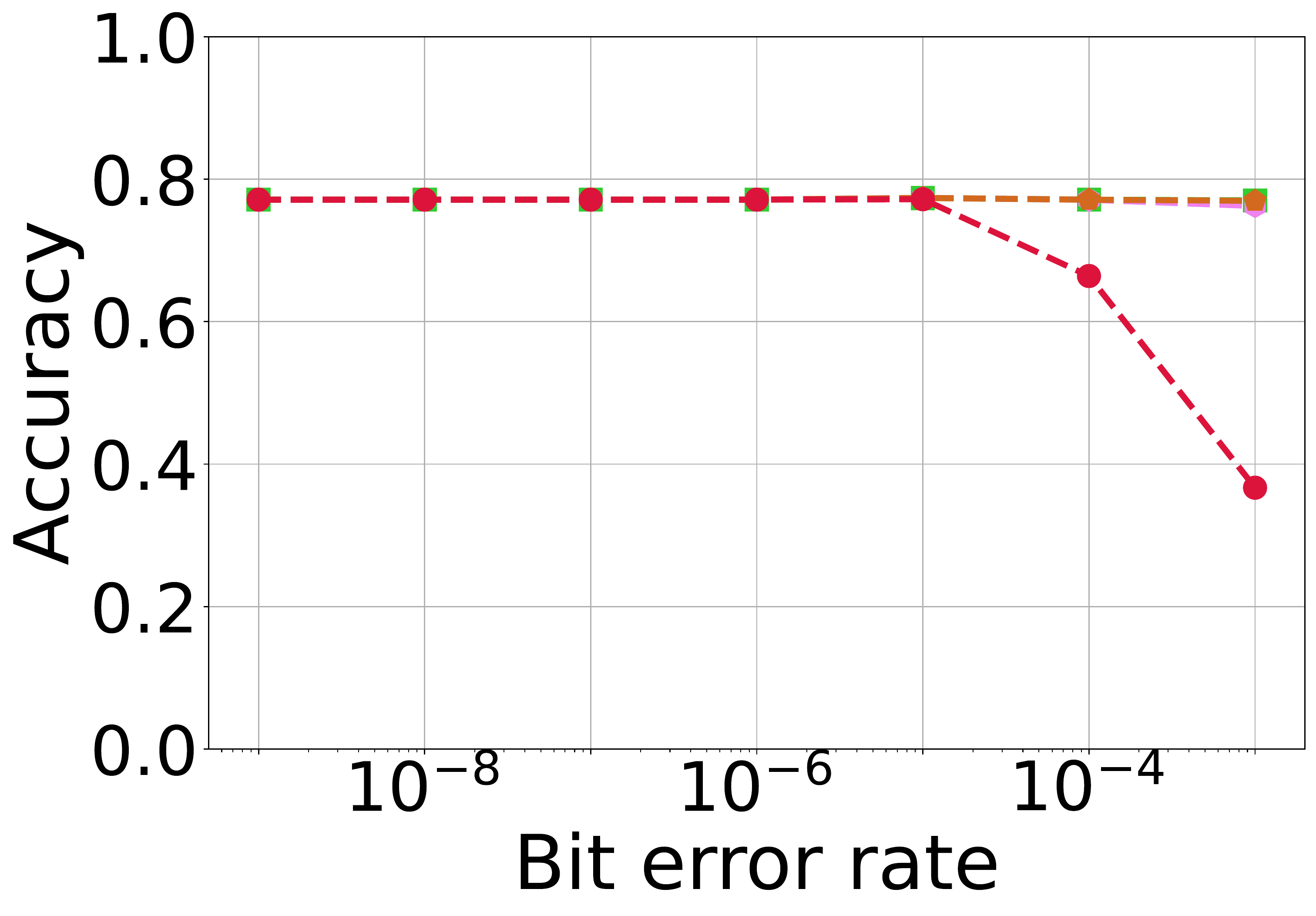}
    }
    \subfigure[GAT]{
        \includegraphics[width=0.473\columnwidth]{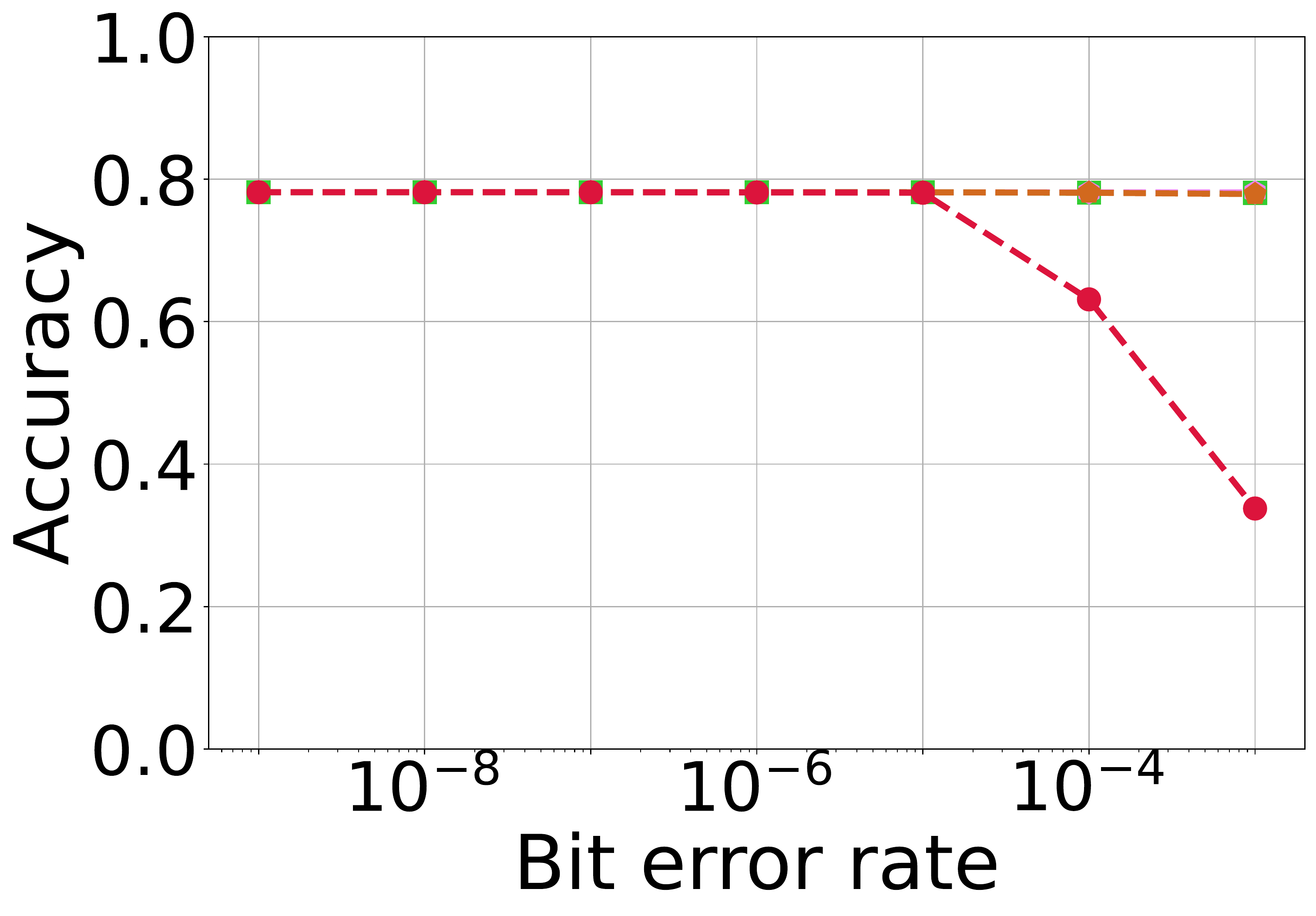}
    }
    \subfigure[Chebyshev]{
        \includegraphics[width=0.473\columnwidth]{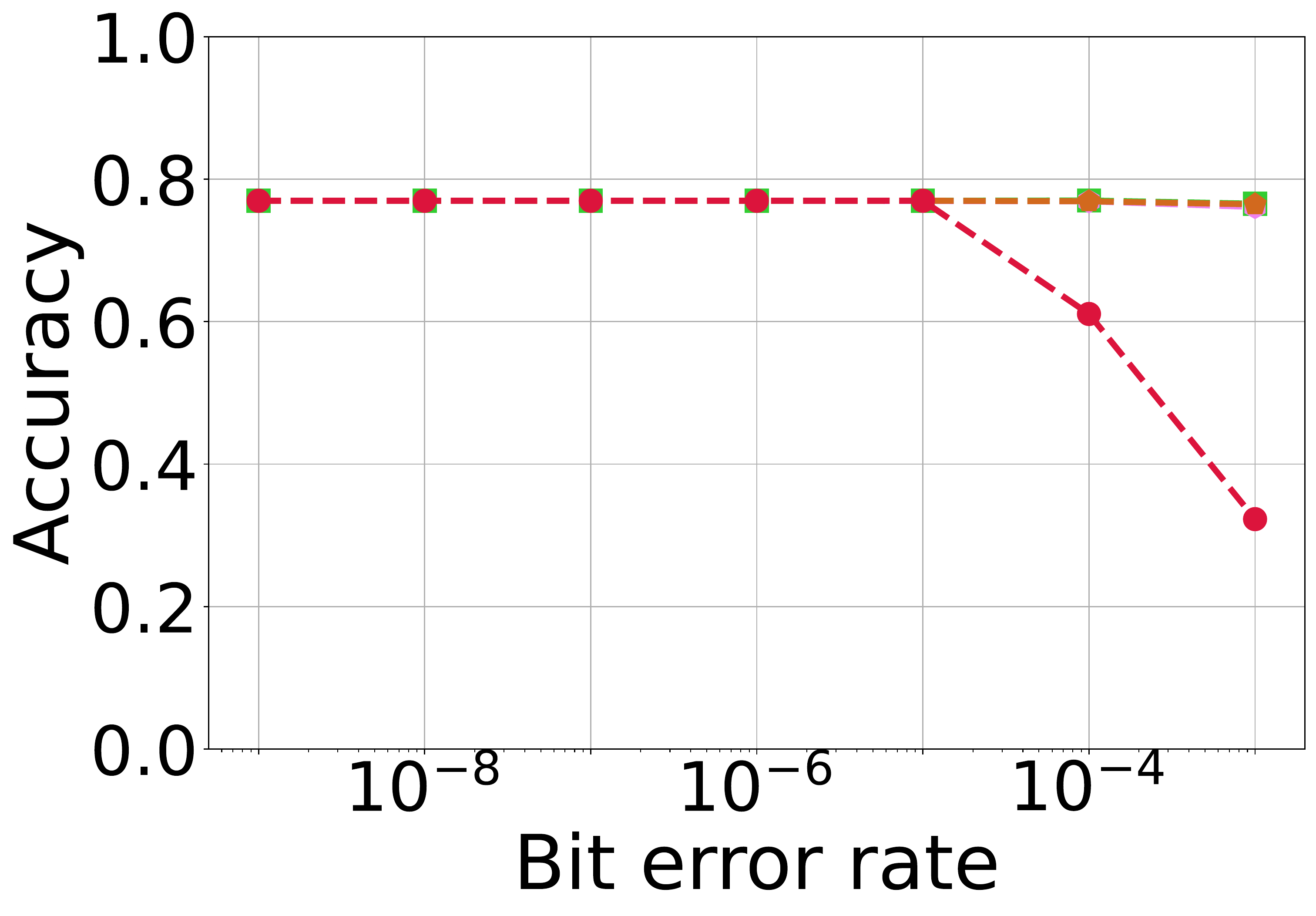}
    }
    \subfigure[SGC]{
        \includegraphics[width=0.473\columnwidth]{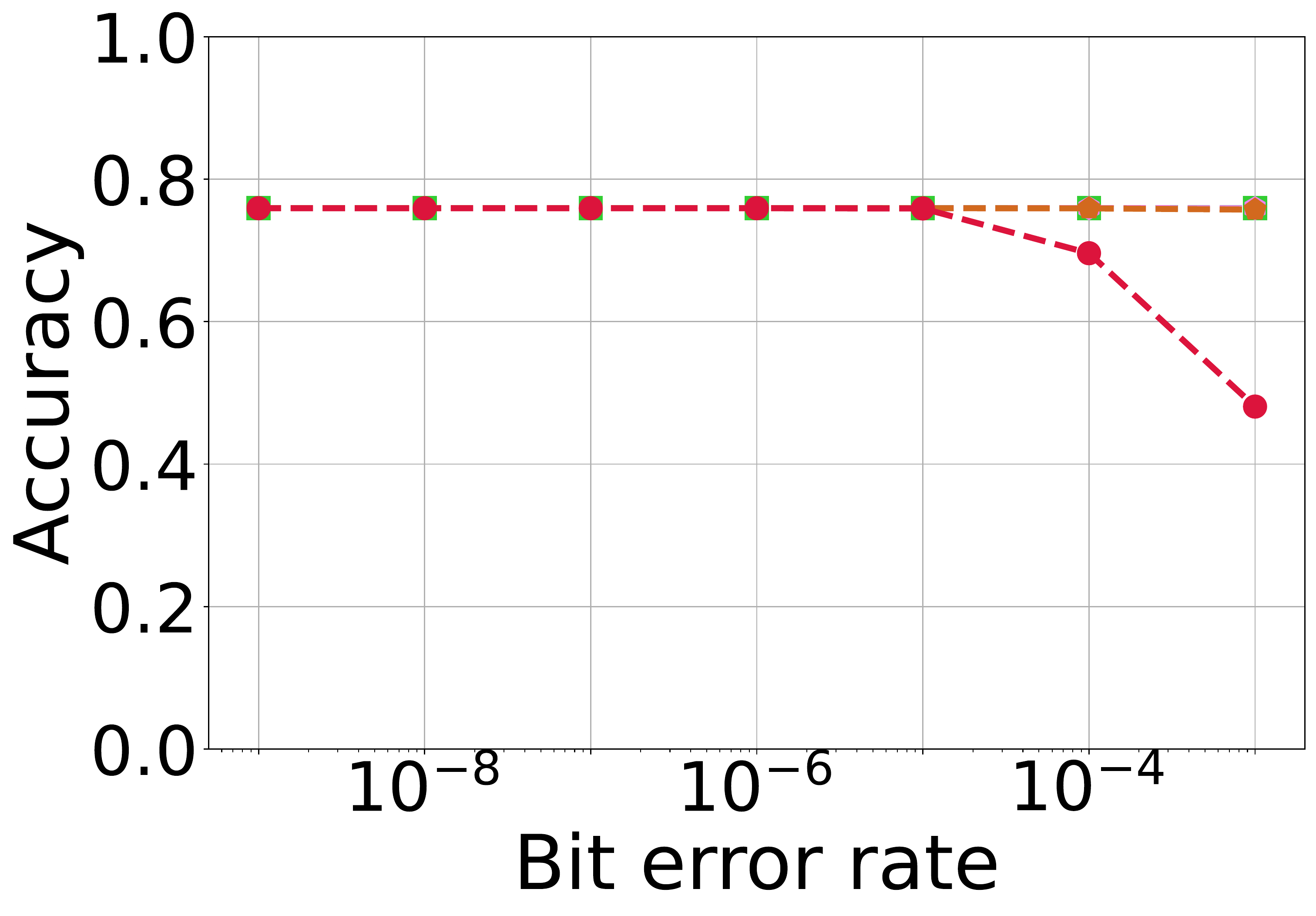}
    }
    \subfigure{
        \includegraphics[width=1.2\columnwidth]{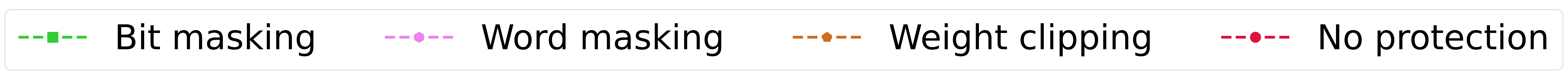}
    }
    \caption{Experimental results for weight error mitigation (averaged over 3 datasets).}
    \label{fig:mitigation1}
    % \vspace{-0.5cm}
\end{figure*}

\begin{figure*}[htbp!]
    \centering

    \subfigure[GCN]{
        \includegraphics[width=0.473\columnwidth]{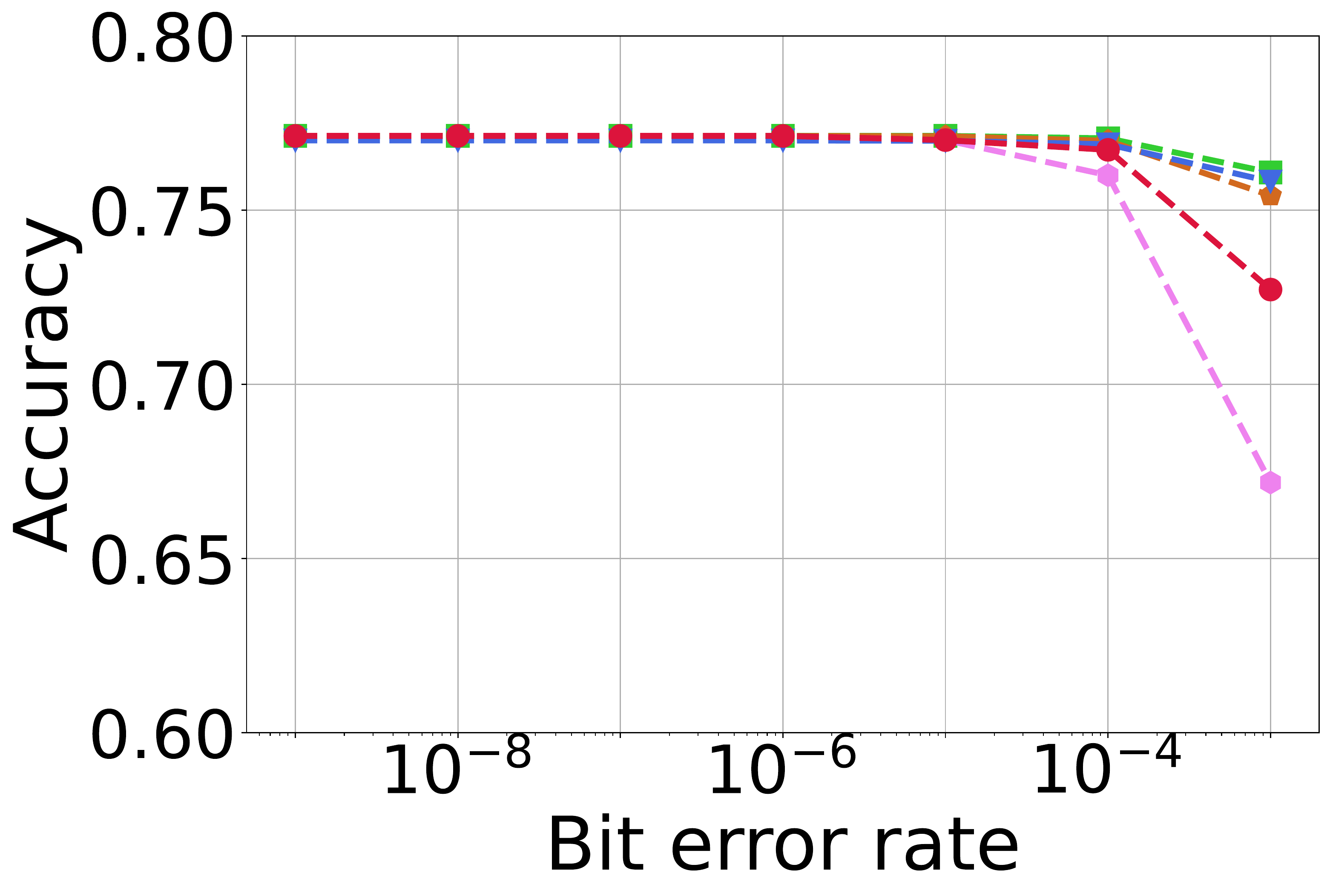}
    }
    \subfigure[GAT]{
        \includegraphics[width=0.473\columnwidth]{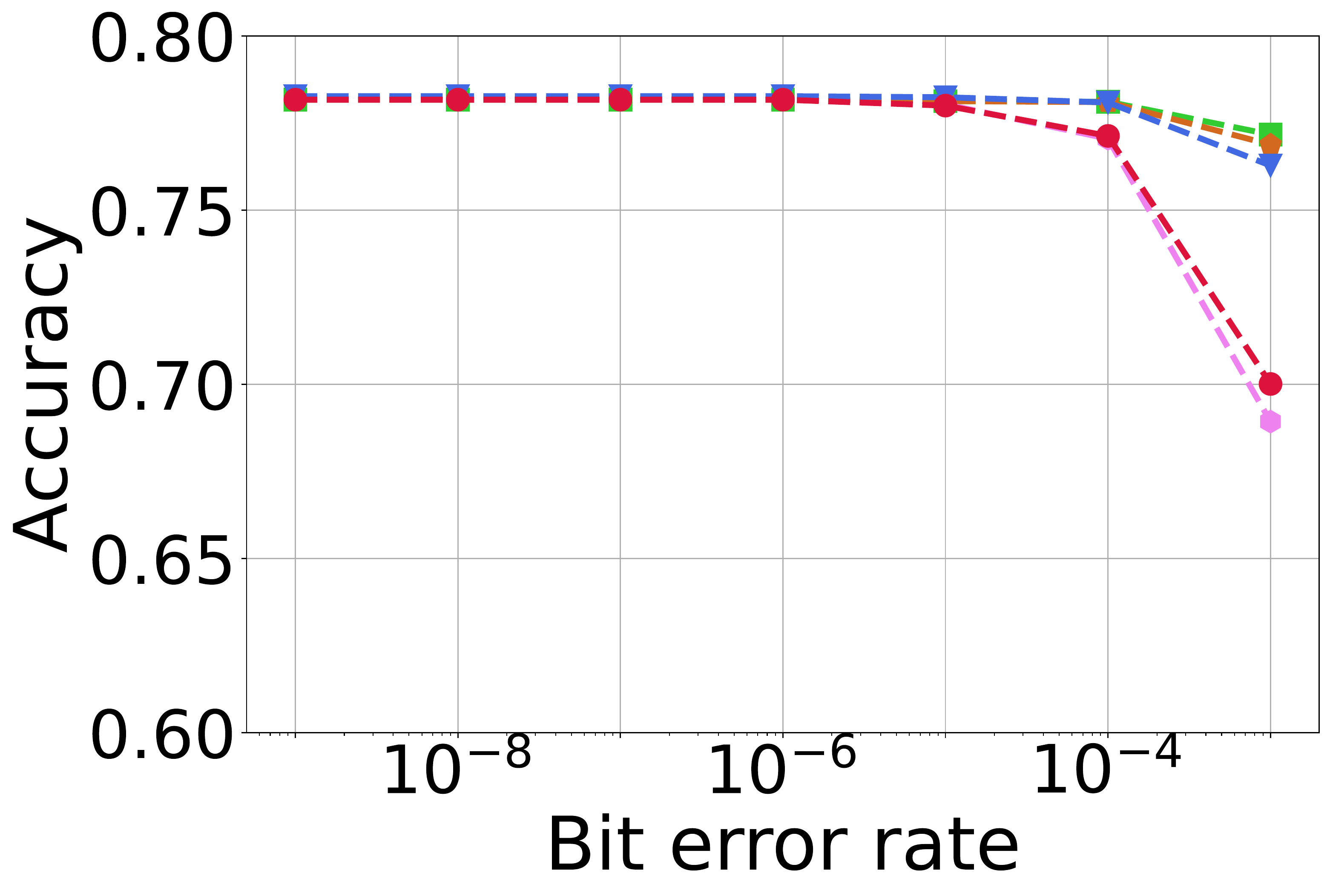}
    }
    \subfigure[Chebyshev]{
        \includegraphics[width=0.473\columnwidth]{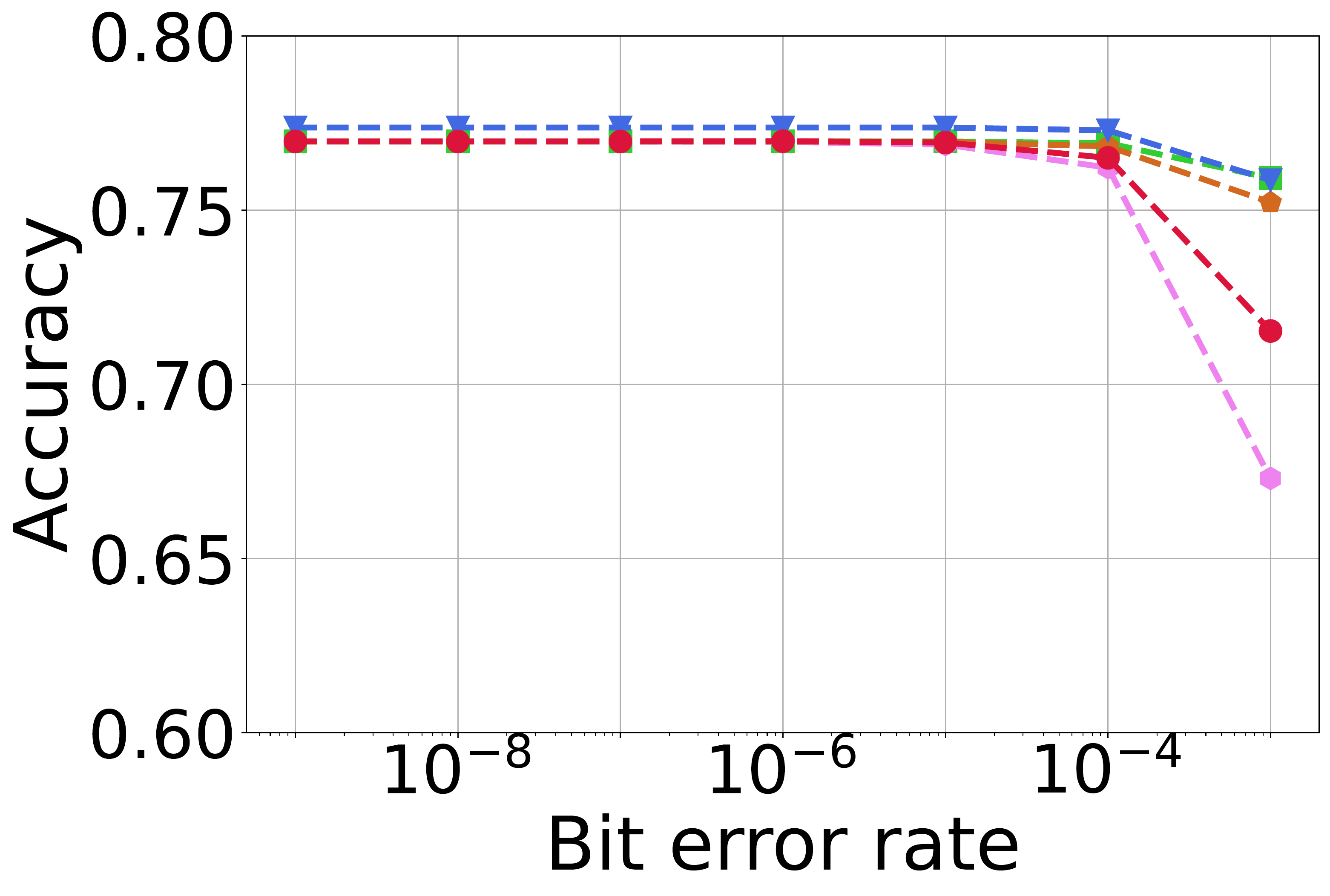}
    }
    \subfigure[SGC]{
        \includegraphics[width=0.473\columnwidth]{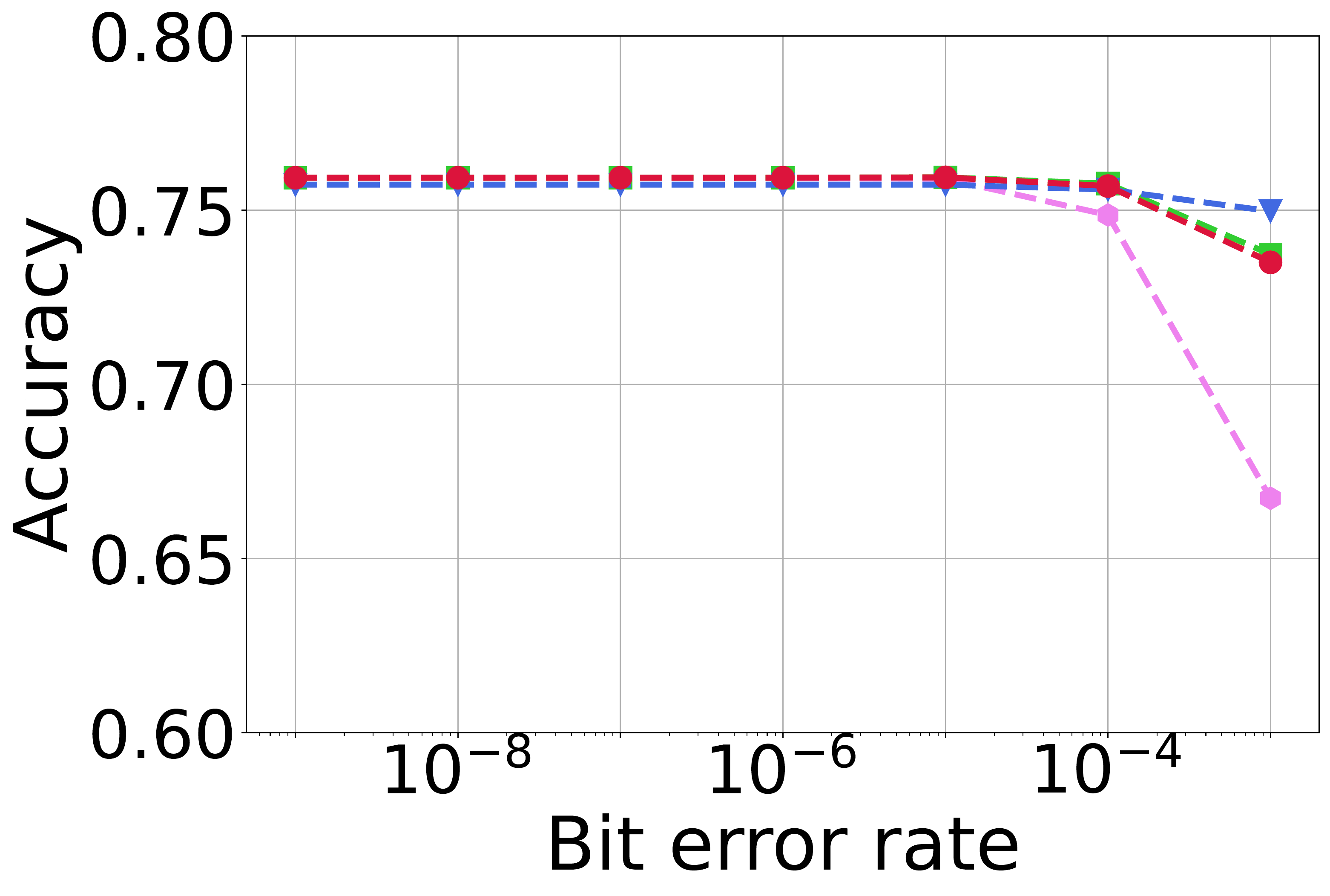}
    }
    \subfigure{
        \includegraphics[width=1.6\columnwidth]{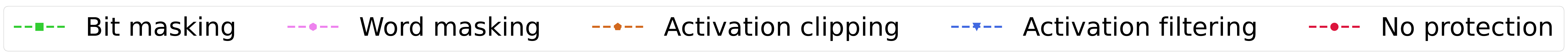}
    }
    % \vspace{-0.4cm}
    \caption{Experimental results for activation error mitigation (averaged over 3 datasets).}
    \label{fig:mitigation2}

\end{figure*}

\subsection{Sensitivity Analysis}

\textbf{Layer Sensitivity}
We further study the sensitivity of different layers in the four evaluated GNN networks as shown in Fig.~\ref{fig:tsd_comp}. Note that SGC only has one GNN layer so this study does not apply to SGC. 
Each of the other three GNNs has two layers, e.g., GCN has GCN-1 and GCN-2, while GAT has GAT-1 and GAT-2. Across all three GNNs, we can see that under the same fault rates, the curve of the first layer always drops much earlier than the second layer with orders of magnitude difference. This sensitivity difference owes to the fact that different layers have an intrinsic difference in their positions of the GNNs and unique characteristics such as the number of parameters, etc. Note that the same phenomenon has been observed in CNNs~\cite{reagen2018ares}, where earlier layers also show much higher sensitivity than later layers.
%This can be explained in multiple ways, where one reason could be that GNN-1 has more parameters than GNN-2. 
%first layer is always more sensitive to the second layer. In other words, 

\textbf{Weights vs. Activation}
We also study the sensitivity difference between weights and activation output (i.e., the intermediate results of GNNs). The difference between them is that weights are read-only while activation outputs are read and then rewritten as the GNNs conduct the operations forward. From Fig.~\ref{fig:tsd_comp}, we can see that activation outputs are less sensitive than weights, which is also consistent with 
previous findings in CNNs~\cite{reagen2018ares}. 

Because the parameters in GNNs are floating point numbers in IEEE-754 standard, a bit-flip can make a number to not-a-number (NAN) case. Thus, we categorize the bit-flips error into four types of erroneous: NaN errors in weights and activation values, and non-NaN errors in weights and activation values. According to our experiment, we find that these four types of error will have different impacts on the accuracy of GNNs models. Generally speaking, NaN error in model weight has the most severe error because it can influence all the nodes during forwarding and leads to a dramatic corruption in the classification accuracy, while the NaN value in activation may only influence the corresponding local node and its neighbors without propagating further. Actually, even non-NaN error in weight can cause more severe consequences than NaN value in activation value due to the same reason. 

%$>$ non-NaN error in weight $>$ NaN error in activation $>$ non-NaN error in activation. According to our observation, in the GNNs model, the NaN value in activation may only influence the corresponding node and its neighbors, which may not catastrophically decrease the overall classification accuracy. However, when the NaN error happens in the weight matrix in the GNNs model, it may influence all the nodes during forwarding and leads to a dramatic corruption in the classification accuracy. On the other hand, comparing the non-NaN errors between weights and activation, the non-NaN errors in activation have a more partial influence on node representations than non-NaN errors in weights.

%One reason to explain this is that activation outputs are usually behind weights in terms of the placement in the GNN networks, since activation output are the results of computations involving weights. Therefore, the activations do not have  the activation output are usually plac

\subsection{Error Mitigation of GNNs}
\textbf{RQ3:} Is there an effective way to mitigate the impact of hardware errors on the GNNs model and to improve the robustness of the GNN system?

\textbf{Ans:} \textbf{Yes, at both hardware and software levels.}

We explore the error mitigation methods described in Section~\ref{sec:mitigation} to mitigate the impact of hardware errors on GNNs models. At the hardware level, we explore two error masking approaches with word-level and bit-level granularity, respectively. For word masking, upon detecting any bit flips, it sets the entire word to 0. While for bit-level masking, it only recovers the faulty bit(s) to 0. We further propose weight and activation clipping and topological-aware activation filtering to mitigate the influence of bit-flip error. The results are shown in Fig.~\ref{fig:mitigation1} and Fig.~\ref{fig:mitigation2}, where Fig.~\ref{fig:mitigation1} focuses on errors occurring in weights and Fig.~\ref{fig:mitigation2} aims at errors in activation output. In this paper, bit-masking and word-masking can mitigate bit-flip errors in both weights and activation outputs, while weight clipping, activation clipping and activation filtering are component-specific error mitigation methods.

According to the experimental results, we can observe several important facts. First of all, for the model weights error scenarios, we can observe that both the masking and clipping methods can provide noticeable error mitigation and significantly enhance the GNNs model robustness. For example, as Fig.~\ref{fig:mitigation1} illustrated, with the bit masking, word masking and weight clipping method, GAT can maintain an acceptable performance under $10^{-3}$ error rate. This is a $1000X$ bit-flip error robustness enhancement compared with the non-protected GAT models which have an accuracy decreasing start with $10^{-6}$ error rate. In fact, we notice that the three approaches have mostly overlapped curves which indicates that the three deliver similar error mitigation. To this end, the weight clipping approach can achieve similar robustness enhancement compared with masking-based methods without hardware modification. Similar error mitigation effects are shown in GCN, CHEB and SGC models. 

Second, for the activation error, activation filtering and bit-masking are noticeably effective to mitigate the bit-flip errors at the activation level, as Fig.~\ref{fig:mitigation2} indicated. Using the GAT model as an instance, the activation filtering, bit-masking and activation clipping approach provide $100X$ error robustness compare with the no-protection case by maintaining the original accuracy at $10^{-5}$ error rate. However, the word masking cannot perform as expected and cannot provide an observable accuracy recovery. For the word-masking approach, we notice that the word masking somehow leads to more accuracy drop than the non-protected scenarios. The reason is that the word masking will overkill the value with bit-flip errors to ``0'' and leads to a misclassification on the corresponding feature lacking node.

Similar error mitigation effects are noticed in GCN and CHEB model experiments. However, for the SGC model, we find that the activation clipping approach shows limited effects. In fact, the activation clipping curve and the no-protection curve are overlapping for the SGC model in Fig.~\ref{fig:mitigation2}. Based on our analysis, errors injected in model activation outputs can directly influence the final classification result, when the bit-flip occurs in the $LogSoftmax$ outputs. Since $LogSoftmax$ is the only activation in the SGC model, the bit-flip errors in the decision-influence outputs are difficult to be corrected by the clipping method we utilized.

\subsection{Implication on Silicon}
\label{sec:discuss}
Strong robustness can be directly translated into improved energy efficiency. For example, if using voltage-scaled on-chip SRAMs fabricated in FDX22 (22nm) technology~\cite{di2019pushing} for GNN weights memory, the inherent GNN robustness (up to $10^{-6}$) can be leveraged to enable voltage scaling corresponding to 42.1\% energy saving, while bit-level masking can further improve the saving to 59.1\%. We also compare the GNN robustness with CNN robustness. We find that (1) similar to CNNs~\cite{reagen2018ares, li2017understanding, mahmoud2020pytorchfi, kim2018matic}, GNNs also exhibit orders of magnitude difference in robustness across different models and application datasets. (2) Many widely-used CNNs and GNNs share similar robustness with similar ``cut-off fault rates'' between $10^{-7}$ and $10^{-5}$~\cite{reagen2018ares, li2017understanding, mahmoud2020pytorchfi, kim2018matic}, at which there is no notable accuracy drop relative to the model’s baseline accuracy. This fault tolerance is significantly higher than conventional computing~\cite{jesd2182010solid}. 
% That is being said, providing same performance, bit masking and word masking strategy can achieve 24.8\% and 13.2\% power saving respectively~\cite{zhang2022energy}. 
%This study can be leveraged to provide insights for efficient GNN system design.  designers during the GNN system design loop. For example, since GNNs are considerably resilient to hardware errors, designers can leverage this robustness to optimize the accuracy-energy-performance design tradeoff. 

\section{Related Work}
With the increasing use of DL methods in safety-critical systems such as robots and healthcare, as well as the pertinent development of energy-efficient DL accelerators, DL robustness (or balancing the robustness-efficiency tradeoff) has been intensively studied in the past, which mainly focuses on the convolutional/recurrent neural networks (CNNs/RNNs)~\cite{reagen2018ares, li2017understanding, mahmoud2020pytorchfi, kim2018matic}. 
Li et al. empirically evaluate the robustness of CNNs at the software level, and found that the robustness of a CNN model depends on data types, values, data reuses, and types of layers in the design~\cite{li2017understanding}.
In the same spirit, \textit{Ares}~\cite{reagen2018ares} proposes a lightweight fault injection framework that can be accelerated on GPUs to explore larger design space. \textit{Ares} examines both the CNN and RNN models with errors in variant models and layers to analyze the influence of hardware errors.
Further, Mahmoud et al. develop a runtime fault injection framework to inject bit-flip errors in both the training and inference process and analyze the error resilience of the computer vision model under different scenarios, including object detection and image classification~\cite{mahmoud2020pytorchfi}. FT-ClipAct~\cite{hoang2020ft} shows that hardware faults that impact the parameters of deep neural networks (DNNs) (e.g., weights) can have drastic impacts on its classification accuracy, and proposes a novel error mitigation technique that squashes the high-intensity faulty activation values to alleviate fault impact. 
\textit{MATIC}~\cite{kim2018matic} searches the impact of memory-related faults and the energy efficiency of DL accelerators according to voltage scaling and enables robust and efficient operation on DNNs accelerators with lower energy consumption.

Given the large body of robustness research on CNNs or RNNs models, to the best of our knowledge, there are no existing studies on GNNs robustness. This paper presents the first effort in this direction by developing an error injection tool for GNNs and assessing its robustness. Our aim is to offer timely insights and early guidance for designing resilient and efficient GNN systems. %We compare CNN and GNN robustness in Section~\ref{sec:discuss}. 

\section{Conclusion}
GNNs are an emerging family of learning methods specialized in learning graph data structures and have been widely used in practical application domains. This paper presents the first effort in systematically assessing and analyzing the robustness of GNNs by developing a GNNs-specific error injection tool on top of PyTorch and performing extensive fault injection experiments across different GNNs models, application datasets, error rates, and GNNs components. Experimental results show that the robustness of GNNs models varies by orders of magnitude with respect to different models and application datasets. We further investigate a low-cost error mitigation mechanism that can improve the error robustness of GNNs models by orders of magnitude. While GNNs are under heavy research in terms of performance and efficiency, this paper aims to shed light on robustness and reliability in GNNs developments.
\bibliographystyle{plain}
\bibliography{main}
\end{document}